%% file: arxiv_gazecap.tex
\documentclass[10pt,twocolumn,letterpaper]{article}

\usepackage{cvpr}
\usepackage{times}
\usepackage{epsfig}
\usepackage{graphicx}
\usepackage{amsmath}
\usepackage{amssymb}
\usepackage{pdfpages}

\usepackage[pagebackref=true,breaklinks=true,letterpaper=true,colorlinks,bookmarks=false]{hyperref}

\include{mymacros}

\usepackage{array}
\usepackage{kotex}
\usepackage{enumerate}
\usepackage{adjustbox}
\usepackage{caption}
\usepackage{subcaption}
\usepackage{color}
\usepackage{slashbox}

\usepackage{booktabs}       %
\usepackage{morefloats}
\usepackage{dblfloatfix}
\usepackage{placeins}

\cvprfinalcopy %

\def\cvprPaperID{156} %

\ifcvprfinal\fi
\begin{document}

\title{Supervising Neural Attention Models for Video Captioning by Human Gaze Data}

\author{
    Youngjae Yu$^\dagger$   \hspace{7pt}
    Jongwook Choi$^\dagger$ \hspace{7pt}
    Yeonhwa Kim$^\ddagger$  \hspace{7pt}
    Kyung Yoo$^\ddagger$    \hspace{7pt}
    Sang-Hun Lee$^\ddagger$ \hspace{7pt}
    Gunhee Kim$^\dagger$ \\
$^\dagger$Department of Computer Science and Engineering \hspace{5pt}
$^\ddagger$Department of Brain \& Cognitive Sciences \\
Seoul National University, Seoul, Korea\\
{\tt\small yj.yu@vision.snu.ac.kr} \hspace{4pt}
{\tt\small \{wookayin,\,billygma,\,yookyung20484,\,visionsl,\,gunhee\}@snu.ac.kr }
}

\maketitle\thispagestyle{empty}

\begin{abstract}
The attention mechanisms in deep neural networks are inspired by human's attention that sequentially focuses on the most relevant parts of the information over time to generate prediction output.
The attention parameters in those models are implicitly trained in an end-to-end manner,
yet there have been few trials to explicitly incorporate human gaze tracking to supervise the attention models.
In this paper, we investigate whether attention models can benefit from explicit human gaze labels, especially for the task of video captioning.
We collect a new dataset called VAS, consisting of movie clips, and corresponding multiple descriptive sentences along with human gaze tracking data.
We propose a video captioning model named \textit{Gaze Encoding Attention Network} (GEAN)
that can leverage gaze tracking information to provide the spatial and temporal attention for sentence generation.
Through evaluation of language similarity metrics and human assessment via Amazon mechanical Turk, 
we demonstrate that spatial attentions guided by human gaze data indeed improve the performance of multiple captioning methods.
Moreover, we show that the proposed approach achieves the state-of-the-art performance for both gaze prediction and video captioning
not only in our VAS dataset but also in standard datasets (\eg  LSMDC~\cite{rohrbach-arxiv-2016} and Hollywood2~\cite{mathe-pami-2015}). 
\end{abstract}

\section{Introduction}
\label{sec:introduction}

Attention-based models have recently gained much interest as a powerful deep neural network architecture
in a variety of applications, including image captioning~\cite{xu-icml-2015}, video captioning~\cite{yao-iccv-2015},
action recognition~\cite{sharma-arxiv-2015}, object recognition~\cite{ba-iclr-2015}, and machine translation~\cite{Bahdanau-iclr-2015} to name a few.
The attention models are loosely inspired by visual attention mechanism of humans,
who do not focus their attention on the entire scene at once,
but instead sequentially adjust the focal points on different parts of the scene over time.

Although the attention models simulate human's attention,
surprisingly there have been few trials to explicitly incorporate human gaze tracking labels to supervise the attention mechanism.
Usually attention models are trained in an end-to-end manner, and thus attention weights are implicitly learned.
In this paper, we aim at investigating whether the explicit human gaze labels can better guide attention models and eventually enhance their prediction performance.
We focus on the task of \textit{video captioning}, whose objective is to generate a descriptive sentence for a given video clip.
We choose the video captioning because the attention mechanism may have more room to play a role in summarizing a sequence of frames that may contain too much information for a short output sentence.
It is worth noting that our objective is not to replace existing video captioning methods for every use case,
given that acquisition of human gaze data is expensive.
Instead, we study the effect of supervision by human gaze for attention mechanism, which has not been discussed in previous literature.

We collect a new dataset named VAS (\textit{Visual Attentive Script}), consisting of movie videos of 15 seconds long,
with multiple descriptive sentences and gaze tracking data.
For pretraining and evaluation of models, we also leverage large-scale caption-only LSMDC dataset~\cite{rohrbach-arxiv-2016}
and gaze-only Hollywood2 eye movement dataset~\cite{marszalek-cvpr-2009,mathe-pami-2015}. %

To explicitly model the gaze prediction for sentence generation,
we propose a novel video captioning model named \textit{Gaze Encoding Attention Network} (GEAN).
The encoder generates pools of visual features depending on not only content and motion in videos,
but also gaze maps predicted by the recurrent gaze prediction (RGP) model. %
The decoder generates word sequences by dynamically focusing on the most relevant subsets of the feature pools. %

Through quantitative evaluation using language metrics and human assessment via Amazon Mechanical Turk (AMT),
we show that human gaze indeed helps enhance the video captioning accuracy of attention models.
One promising result is that our model learns from a relatively small amount of gaze data of VAS and Hollywood2 datasets,
and improves the captioning quality on LSMDC dataset with no gaze annotation.
It hints that potentially we could leverage gaze information in a semi-supervised manner, and apply domain adaptation or transfer learning to boost the performance further.

To conclude the introduction, we highlight major contributions of this work as follows.

(1) To the best of our knowledge, our work is the first to study the effect of supervision by human gaze data on attention mechanisms, especially for the task of video captioning.
We empirically show that the performance of multiple video captioning methods increases with the spatial attention learned from human gaze tracking data.

(2) We collect the dataset called VAS, consisting of 15 second-long movie clips, and corresponding multiple descriptive sentences and human gaze tracking labels.
As far as we know, there has been no video dataset that associates with both caption and gaze information.

(3) We propose a novel video captioning model named \textit{Gaze Encoding Attention Network} (GEAN) that efficiently incorporates spatial attention
by the gaze prediction model with temporal attention in the language decoder.
We demonstrate that the GEAN achieves the state-of-the-art performance for both gaze prediction and video captioning
not only in our VAS dataset but also in the standard datasets (\eg  LSMDC~\cite{rohrbach-arxiv-2016} and Hollywood2~\cite{mathe-pami-2015}).

\textbf{Related work}.
We briefly review several representative papers of video captioning.
Although several early models successfully tackle the video captioning based on the framework of
CRF~\cite{rohrbach-iccv-2013}, topic models~\cite{das-cvpr-2013}, and hierarchical semantic models~\cite{guadarrama-iccv-2013},
recent advances in deep neural models have led substantial progress for video captioning.
Especially, multi-modal recurrent neural network models have been exploited as a dominant approach;
some notable examples include \cite{jeff-cvpr-2015,rohrbach-gcpr-2015,venugopalan-iccv-2015,venugopalan-hlt-2015}.
These models adopt encoder-decoder architecture;
the encoder represents the visual content of video input via convolutional neural networks,
and the decoder generates a sequence of words from the encoded visual summary via recurrent neural networks.
Among papers in this group,
\cite{yao-iccv-2015} and \cite{yu-cvpr-2016} may be the most closely related to ours,
because they are also based on attention mechanisms for caption generation.
Compared to all the previous video captioning methods, %
the novelty of our work is to leverage the supervision of attention using human gaze tracking labels.
Moreover, our experiments show that such gaze information indeed helps improve video captioning performance.

\begin{table*}[t]
\centering
\setlength{\tabcolsep}{4pt}
\begin{tabular}{l|cccccc}
\hline
                                                         & \multirow{2}*{\# videos} & \# sentences & Vocabulary & Median length & \# gaze data & \multirow{2}*{\# subjects} \\
                                                         &                          & (per video)  & size       & of sentence   & (per video)  &                            \\ \hline
VAS                                                      & 144                      & 4,032 (28)   & 2,515      & 10            & 1,488 (10--11) & 31                         \\ \hline
LSMDC~\cite{rohrbach-arxiv-2016}                         & 108,470                  & 108,536 (1--2)  & 22,898     & 6             & --           & --                         \\ \hline
Hollywood2 EM~\cite{marszalek-cvpr-2009,mathe-pami-2015} & 1,707                    & --           & --         & --            & 27,312 (16)  & 16                         \\
\hline
\end{tabular}
\vspace{-3pt}
\caption{Statistics of our novel VAS, caption-only LSMDC, and gaze-only Hollywood2 EM datasets.}
\label{tbl:dataset}
\end{table*}

\section{Video Datasets for Caption and Gaze}
\label{sec:dataset}

We use three movie video datasets, including (i) caption-only LSMDC~\cite{rohrbach-arxiv-2016}, and (ii) gaze-only Hollywood2 EM (Eye Movement)~\cite{marszalek-cvpr-2009,mathe-pami-2015},
and (iii) our newly collected VAS dataset with both captions and gaze tracking data.
Since the LSMDC and Hollywood2 EM are more large-scale than our VAS, they are jointly leveraged for pretraining.
Table \ref{tbl:dataset} summarizes some of basic statistics of the datasets.

\textbf{LSMDC}~\cite{rohrbach-arxiv-2016}. This dataset is a combination of recently published two large-scale movie datasets, %
MPII-MD \cite{rohrbach-cvpr-2015}  and M-VAD \cite{torabi-mvad-2015}. %
It consists of 108,470 clips in total,
and associates about one sentence with each clip. The text is obtained from the descriptive video service (DVS) of the movies.
The clips of MPII-MD and and M-VAD are sampled from 72 and 92 commercial movies, and have lengths of 3.02 and 6.13 seconds long on average, respectively.

\textbf{Hollywood2 EM}~\cite{marszalek-cvpr-2009}. This dataset is originally proposed for action recognition of 12 categories from 69 movies.
Later \cite{mathe-pami-2015} collects eye gaze data from 16 subjects for all 1,707 video clips, using the SMI iView X HiSpeed 1250 eye tracker.

\textbf{VAS}. The \textit{Visual Attentive Script} (VAS) dataset
includes 144 emotion-eliciting clips of 15 seconds long. %
For each video clip, we collect multiple tracking data of subjects' gazes and pupil sizes using EyeLink 1000 plus eye tracker.
We invite 31 subjects, each of whom generates eye gaze data for 48 clips.
We let subjects to freely watch a video clip to record gaze tracking, and then request to describe it in three different sentences
(\ie one general summary sentence, and two focused sentences on storyline, and characters on background).
Since clips are sampled from commercial movies, we observe rather stable gaze tracking across subjects.
Also, a 15-sec clip often includes much content; it can be easier for subjects to resolve their understanding with different aspects of short sentences.
We defer the details of data collection and analyses to the supplementary.

\section{Approach}
\label{sec:approach}

We propose \textit{Gaze Encoding Attention Networks} (GEAN), as shown in Fig.\ref{fig:archi}. %
We first extract three types of CNN features for scene, motion, and fovea per frame (section \ref{sec:viddesc}).
The recurrent gaze prediction (RGP) model learns from human gaze to decide which parts of scenes to be focused (section \ref{subsec:gaze_prediction}).
The encoder creates feature pools using content and motion in a video with spatial attention guided by the RGP model (section \ref{subsec:feature_pool}).
The decoder produces a word sequence by sequentially focusing on the most relevant subsets of the feature pools (section \ref{subsec:capgen_model}).

\subsection{Video Pre-processing and Description}
\label{sec:viddesc}

We equidistantly sample one per five frames from a video, to reduce the frame redundancy and memory consumption while minimizing loss of information.
We denote the number of video frames as $N$.
We extract three types of video features (\ie \textit{scene}, \textit{motion}, and \textit{fovea} features), all of which have dimensions of $1,024$.
(1) \textbf{Scene}: To present a holistic view of each video scene, we extract the scene description from the pool5/7{\sf x}7s1 layer of GoogLeNet~\cite{szegedy-cvpr-2015} %
that is pretrained on Places205 \cite{zhou-nips-2014} dataset.
Each input frame is scaled to $256\times256$, and center-cropped to a $227\times227$ region.
(2) \textbf{Motion}: We extract spatio-temporal motion representation from the conv5b layer (\ie $\mathbb R^{7 \times 7 \times 1,024}$)
of the pretrained C3D network~\cite{tran-iccv-2015} on Sports-1M dataset \cite{karpathy-cvpr-2014}.
For each frame, we input a sequence of previous 16 frames to the C3D.
The input frames are scaled to $112\times112$. %
(3) \textbf{Fovea}:
We extract the frame representation from the inception5b layer (\ie $\mathbb R^{7 \times 7 \times 1,024}$) of GoogLeNet~\cite{szegedy-cvpr-2015} pretrained on ImageNet dataset~\cite{imagenet-ijcv-2015},
which is later weighted by spatial attention.
The input frames are scaled to $227\times227$ without center-cropping to ensure that peripheral regions are not cropped out.
We defer the details of how the spatial attention weights on these features to section \ref{subsec:feature_pool}.

To build a dictionary, we first tokenize all words except punctuation from LSMDC and VAS datasets, using {\small\tt wordpunct$\_$tokenizer} of the NLTK toolbox~\cite{nltk}.
We perform lowercasing and retain rare words to reserve the originality of caption datasets.
In captions, we replace proper nouns like characters' names by \textit{SOMEONE} token.

\subsection{The Recurrent Gaze Prediction (RGP) Model}
\label{subsec:gaze_prediction}

\begin{figure*}[t]
\centering
\includegraphics[width=0.95\textwidth]{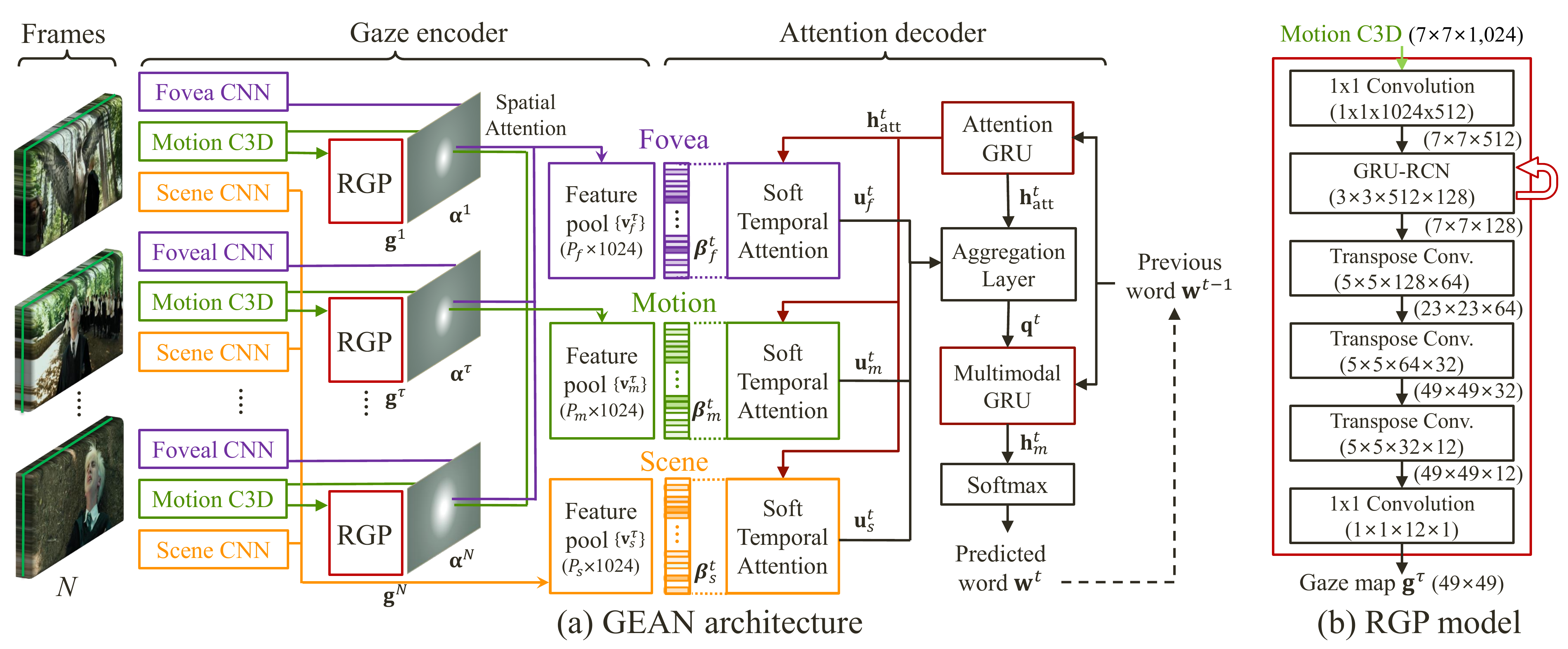}
\vspace{-6pt}
\caption{\small
Illustration of the proposed \textit{Gaze Encoding Attention Network} (GEAN) architecture.
After extracting three types of CNN features per frame,
the encoder generates pools of visual features using content and motion in videos, weighted by the spatial attention that the recurrent gaze prediction (RGP) model generates. %
The decoder generates word sequences by sequentially focusing on the most relevant subsets of the feature pools.
In the RGP model, we present filter sizes inside boxes, and output dimensions next to arrows.
}
\label{fig:archi}
\vspace{-6pt}
\end{figure*}

The goal of the RGP model is to predict a gaze map per frame of an input video, after learning from human gaze tracking data.
The output gaze map $\mathbf g^\tau$ at frame $\tau$ is defined as a $\ell_1$-normalized ($49 \times 49$) matrix %
that indicates a probability distribution of where to attend in a $49 \times 49$ grid.
We design the RGP model built upon GRUs (Gated Recurrent Units)~\cite{ballas-iclr-2016,cho-emnlp-2014},
followed by three layers of convolution transpose (\ie deconvolution), a $1\times 1$ convolution, and an average-pooling layer.
Fig.\ref{fig:archi}(b) shows the structure.
We choose GRUs since they are empirically superior to model long-term temporal dependency with less parameters.
Since we deal with a frame sequence, we use a variant of GRUs (\ie GRU-RCN in \cite{ballas-iclr-2016}),
which replaces fully-connected units in the GRU with convolution operations:
{
\begin{align}
\label{eq:grurcn}
    \mathbf z^{\tau}          & = \sigma (\mathbf W_{z} * \mathbf x^{\tau} + \mathbf U_{z} * \mathbf h^{\tau-1}), \\
    \mathbf r^{\tau}          & = \sigma (\mathbf W_{r} * \mathbf x^{\tau} + \mathbf U_{r} * \mathbf h^{\tau-1}),   \\
    \tilde{\mathbf h}^{\tau}  & = \mathrm{tanh}(\mathbf W * \mathbf x^{\tau} + \mathbf U * (\mathbf r^{\tau} \odot \mathbf h^{\tau-1})),  \\
    \mathbf h^{\tau}          & = (1-\mathbf z^{\tau}) \mathbf h^{\tau-1} + \mathbf z^{\tau} \tilde{\mathbf h}^{\tau},
\end{align}%
}%
where $\sigma$ is the sigmoid function, $*$ denotes a convolution, and $\odot$ is an element-wise multiplication.
The input $\mathbf x^{\tau}$ at frame ${\tau}$ is the C3D motion feature discussed in section \ref{sec:viddesc},
projected to $(7\times7\times512)$ by a linear transformation (\ie $1\times1$ convolution). %
$\mathbf h^{\tau}$, $\mathbf z^{\tau}$, and $\mathbf r^{\tau}$ denote the hidden state, update gate, and reset gate at $\tau$, respectively,
whose dimensions are all $(7\times7\times 128)$.
Model parameters $\mathbf W_*$ and $\mathbf U_*$ are 2D-convolutional kernels with a size of $k_1 \times k_2 \times O_x \times O_y$,
where $k_1 \times k_2$ is the convolutional kernel size, and $O_x$ and $O_y$ are input and output channel dimensionality.
We set $k_1 = k_2  =3$ as a kernel size.
By using $k_1 \times k_2$ spatial kernels,
the gates $\tilde{\mathbf h}^{\tau}(i,j) , \mathbf z^{\tau}(i,j)$, and $\mathbf r^{\tau}(i,j)$ at location $(i,j)$ depend on both local neighborhood of input $\mathbf x^{\tau}$ and the previous hidden state map $\mathbf h^{\tau-1}$.
Thus, the hidden recurrent representation $\mathbf h^{\tau}$ can fuse a history of 3D convolutional motion features through time while keeping spatial locality.
We then apply a sequence of three transposed convolutions,
followed by another $1\times1$ convolution, and softmax to $\mathbf h^{\tau}$,
to obtain a predicted gaze map $\mathbf g^{\tau}$ of shape $(49 \times 49)$.
Fig.\ref{fig:archi}(b) also presents dimensions and filter sizes for each layer operation.

\subsection{Construction of Visual Feature Pools}
\label{subsec:feature_pool}

We construct three types of feature pools using the features of scene, motion, and fovea discussed in section \ref{sec:viddesc}.
The first feature pool denoted by $\{\mathbf v_s^{\tau}\}_{\tau=1}^N$ is a simple collection of scene features for each frame, where $\tau$ is the frame index from 1 to $N$.
For the next two feature pools, we use the predicted gaze map as spatial attention weights.
Its underlying rationale is that %
human perceives focused regions in a high visual acuity with more neurons, %
while peripheral scene fields in a low resolution with less neurons~\cite{larson-jv-2009}.
Roughly simulating such a mechanism occurring in a focused foveal zone in human's retina,
we obtain a spatial attention map $\bm \alpha^\tau \in \mathbb{R}^{7\times 7}$ by average-pooling $\mathbf g^\tau \in \mathbb{R}^{49\times 49}$ with a $(7\times7)$ kernel,
and adding  a uniform distribution with a strength of $\lambda$.
Our empirical finding is that adding a uniform distribution leads to better performance;
relying on only a very focused region can be risky to ignore too much relevant parts in the scene.
We use $\lambda = 0.6$ via cross validation.
Finally, we $\ell_1$-normalize $\bm \alpha^\tau$ to yield a probability map.
Next we define the motion and fovea feature pools (\ie $\{\mathbf v_m^{\tau}\}_{\tau=1}^{N}$ and $\{\mathbf v_f^{\tau}\}_{\tau=1}^{N}$) as follows.
We compute each $\mathbf v_m^{\tau}$ / $\mathbf v_f^{\tau}$ at frame $\tau$ as a weighted sum of element-wise dot-product between $\bm \alpha^\tau$
and the motion/fovea features, both of which have dimension of $(7 \times 7 \times 1,024)$ as presented in section \ref{sec:viddesc}.
For example, each $\mathbf v_m^{\tau} \in \mathbb R^{1,024}$ is computed as $\mathbf v_m^{\tau} (k) = \sum_{i=1}^7 \sum_{j=1}^7 \bm \alpha^\tau(i,j) \mathbf f_m^\tau(i,j,k)$,
where $\mathbf f_m^\tau$ is the C3D conv5b motion feature at frame $\tau$.

We then set the maximum lengths of pools denoted by $N_{max}$ for scene, motion, and attention features to 20, 35, and 35 respectively, based on the average length of video clips.
If $N < N_{max}$, we repeat padding again from the feature of the first frame; otherwise, we uniformly sample frames to be fit to the limit length.
We use a smaller pool size $P$ for the scene, because its variation across a clip is smaller than other feature types.
We remind that all pooled features have a dimension of $1,024$.

\subsection{The Decoder for Caption Generation}
\label{subsec:capgen_model}

Our decoder for caption generation is designed based on the soft attention mechanism~\cite{Bahdanau-iclr-2015}, which has been also applied in video captioning applications
(\eg\cite{yao-iccv-2015,yu-cvpr-2016}).
Thus, the decoder sequentially generates words by selectively weighting on different features in the three pools at each time.
As shown in Fig.\ref{fig:archi}, the decoder consists of a temporal attention module, an attention GRU, an aggregation layer, and a multimodal GRU.

\textbf{Temporal attention module}.
For each feature pool $\{\mathbf v^{\tau} \}_{\tau}$,
we compute a set of attention weights $\{ \{\beta_\tau^1\}_\tau, \cdots, \{\beta^L_\tau\}_\tau \}$
such that $\sum_{\tau=1}^N \beta_{\tau}^t = 1$ at each time step $t$,
where $N$ is the length of each visual pool, and $L$ is the output sentence length.
Here $t$ indicates the step for a output word sequence; it is different with $\tau$ in the previous section,
which means the frame index.
Thus for each word $t$, the distribution $\{\beta_\tau^t \}_{\tau}$ determines the temporal attention.
Since we have three sets of visual pools $\{\mathbf v_{s,m,f}^{\tau}\}_\tau$, we also have three sets of attention weights $\{\beta\}$.
We let the attention mechanism for each pool to be independent; we below drop the subscript $s,m,f$ for simplicity.
We compute a single aggregated feature vector $\mathbf u^t \in \mathbb R^{1,024}$ by $\beta$-weighted averaging on all the features $\{\mathbf v^{\tau}\}_\tau$ in each pool:
\begin{align}
\label{eq:attent_featvec}
\mathbf u^t &= \sum^{N}_{\tau=1} \beta^{t}_{\tau} \mathbf v^{\tau},
 \hspace{6pt} \mbox{ where }
\beta^{t}_\tau = \frac{ \exp(q^{t}_\tau)} {\sum_{\tau'} \exp(q^{t}_{\tau'})}, \hspace{6pt} \\
    q^t_{\tau} &= \mathbf w^{\top} \phi (\mathbf W_q \mathbf v^\tau + \mathbf U_q \mathbf h_{att}^{t} + \mathbf b_q),
\end{align}%
where each attention weight $\beta^{t}_{\tau}$ is obtained by applying a sequential softmax to scalar attention scores $\{q_{\tau}^t\}_\tau$.
The parameters includes $\mathbf w \in \mathbb R^{64\times1}$, $\mathbf W_q \in\mathbb R^{64\times 1,024}$,
$\mathbf U_q \in\mathbb R^{64\times 512}$
are shared for each feature pool at all time steps.
The activation $\phi$ is a scaled hyperbolic tangent function (\ie $\mathrm{stanh}(x) = 1.7159 \cdot \mathrm{tanh}( 2x/3 )$),
and $\mathbf h_{att}^{t-1} \in \mathbb R^{512}$ is the previous hidden state of the attention GRU, which will be discussed below.

\textbf{Attention GRU}.
Our attention GRU has the same form with the normal GRU~\cite{cho-emnlp-2014} as follows:

{
\vspace{-12pt}
\begin{align}
\label{eq:grurcn}
    \mathbf z_{att}^{t}         &= \sigma (\mathbf W_{z}  \mathbf x_{att}^{t} + \mathbf U_{z}  \mathbf h_{att}^{t-1} + \mathbf b_{z}),   \\
    \mathbf r_{att}^{t}         &= \sigma (\mathbf W_{r}  \mathbf x_{att}^{t} + \mathbf U_{r}  \mathbf h_{att}^{t-1} + \mathbf b_{r} ),    \\
    \tilde{\mathbf h}_{att}^{t} &= \mathrm{tanh}(\mathbf W_{h}  \mathbf x_{att}^{t} + \mathbf U_{h} * (\mathbf r_{att}^{t} \odot \mathbf h_{att}^{t-1})), \\
    \mathbf h_{att}^{t}         &= (1-\mathbf z_{att}^{t})  \odot \mathbf h_{att}^{t-1} + \mathbf z_{att}^{t}  \odot \tilde{\mathbf h}_{att}^{t}.
\end{align}%
}%
The input $\mathbf x_{att}^t$ is an embedding of the previous word:
$\mathbf x_{att}^t = \mathbf B \mathbf w^{t-1}$, where $\mathbf w^{t-1}$ is a $V \times 1$ one-hot vector, and $\mathbf B \in \mathbb R^{512 \times V}$ is a word embedding parameter.
The hidden state representation $\mathbf  h_{att}^{t} $ is the input to both the temporal attention module and the aggregation layer;
that is, it influences not only the attention on the feature pools but also the generation of a next probable word.

\textbf{Aggregation layer}.
Note that the attention feature vectors in Eq.(\ref{eq:attent_featvec}) are obtained for each channel of scene, motion, and fovea separately:
$\mathbf u_s^t$, $\mathbf u_m^t$, and $\mathbf u_f^t$,
which are then fed into the aggregation layer.
{%
\begin{align}
    \label{eq:multimodal_layer}
    \mathbf q^t = \phi ( ([\mathbf W_g^s \mathbf u_s^{t} \parallel \mathbf W_g^m \mathbf u_m^{t} \parallel \mathbf W_g^f \mathbf u_f^{t}]  + \mathbf b_{g}) \odot \mathbf U_{g} \mathbf h_{att}^{t} )
\end{align}
}%
where $\parallel$ denotes the vector concatenation, and parameters include
$\mathbf W_g^s \in \mathbb R^{256 \times 1,024},\mathbf W_g^m \in \mathbb R^{256 \times 1,024}, \mathbf W_g^f  \in \mathbb R^{512 \times 1,024} , \mathbf b_{m} \in \mathbb R^{1,024\times 1}$
and $\mathbf U_{g} \in \mathbb R^{1,024 \times 512}$.
We apply a dropout regularization \cite{JMLR:v15:srivastava14a} with a rate of 0.5 to the aggregation layer,
which mixes each feature channel representation with previous word information via the hidden state $\mathbf h_{att}^t$ of the attention GRU.
It then outputs a vector $\mathbf q^t \in \mathbb R^{1,024}$, based on which the multimodal GRU generates a next likely word.

\textbf{Multimodal GRU}.
The multimodal GRU has the same structure with the attention GRU with only difference that
input $\mathbf x_{m}^t$ is a concatenation of the output of the aggregation layer and the previous word embedding:
$[\mathbf q^t, \mathbf B \mathbf w^{t-1} ] \in \mathbb R^{1,536}$.
That is, the multimodal GRU couples attended visual features with embedding of the previous word.
The hidden state $\mathbf h_{m}^t$ is fed into a softmax layer over all the words in the dictionary to predict the index of a next word:
{
\begin{align}
    \label{eq:output_layer}
    p(\mathbf w^t \mid \mathbf w^{1:t-1}) = \mathrm{softmax}(\mathbf W_{out} \mathbf h_{m}^{t} + \mathbf b_{h}),
 \end{align}
}%
\noindent where parameters include $\mathbf W_{out} \in \mathbb R^{V\times 512 }$ and $\mathbf b_{h} \in \mathbb R^{V \times 1}$.
We use a greedy decoding scheme to choose the best word $\mathbf w^t$ that maximizes Eq.(\ref{eq:output_layer}) at each time step.

\textbf{Spatial and temporal Attention}.
The proposed GEAN model leverages both spatial and temporal attention.
The spatial attention is used for generating feature pools that are weighted by gaze maps predicted by the RGP model.
The temporal attention is used for selecting a subset of feature pools for word generation by modules in the decoder.
By sequentially running the two attentions, we can significantly reduce the dimensionality of spatio-temporal attention compared to other previous work (\eg \cite{sharma-arxiv-2015,yu-cvpr-2016}),
which allows us to train the model with fewer training data.
Moreover, it also resembles human’s perceptual process that is initially sensitive to visual stimuli, and then creates words using the memory about visual experience.

\subsection{Training}
\label{subsec:training}

We first train %
the RGP model, and %
then learn the entire GEAN model while fixing parameters of the RGP model.
This two-step learning leads to better performance than allowing parameter update.

\textbf{Training of the RGP model}.
We obtain groundtruths of gaze maps from human gaze tracking data in the training sets of VAS and Hollywood2.
Following \cite{mathe-pami-2015},
we first build a ($49 \times 49$) binary fixation map from raw gaze data,
and then apply Gaussian filtering with $\sigma = 2.0$ and $\ell_1$-normalization to obtain a ($49 \times 49$) groundtruth gaze map,
which can be seen as a valid probability distribution of eye fixation. %
We use the averaged frame-wise cross-entropy loss between predicted and GT gaze maps.
We minimize the loss with Adam optimizer \cite{kingma-iclr-2015}, with an initial learning rate of $10^{-4}$.
To reduce overfitting further, we use data augmentation of image mirroring. %

\textbf{Training of the GEAN model}.
We limit the maximum length $L$ of training sentences to 80 words.
We use the cross-entropy loss between predicted and GT words with $\ell_2$-regularization to avoid overfitting.
We use orthogonal random initialization for two GRUs, and Xavier initialization~\cite{glorot-AISTATS-2010} for convolutional and embedding layers.
We use Adam optimizer~\cite{kingma-iclr-2015} with an initial learning rate of $10^{-4}$.

\section{Experiments}
\label{sec:experiments}

We first validate the performance of the recurrent gaze prediction (RGP)  model for gaze prediction in section \ref{subsec:results_gazepred}
We then report quantitative results of human gaze supervision on the attention-based captioning in section \ref{subsec:results_vidcap}.
Finally, we present AMT-based human assessment results for captioning quality in section \ref{subsec:results_human}.
We defer more thorough experimental results to the supplementary. %
We plan to make public our source code and VAS dataset. %

For evaluation, we randomly split VAS dataset into 60/40\% as training and test sets.
For LSMDC and Hollywood2 dataset, we use the split provided by original papers \cite{rohrbach-arxiv-2016} and \cite{mathe-pami-2015}, respectively.

\subsection{Evaluation of Gaze Prediction}
\label{subsec:results_gazepred}

\newcolumntype{P}[1]{>{\centering\arraybackslash}p{#1}}
\begin{table*}[t]
\centering
\small
\begin{tabular}{l|P{.6cm}P{.6cm}P{.6cm}P{.6cm}|P{.6cm}P{.6cm}P{.6cm}P{.6cm}}
\hline

\small
                                             & \multicolumn{4}{c|}{VAS}                                  & \multicolumn{4}{c}{Hollywood2 EM} \\ \hline
Metrics                           & Sim   & CC    & sAUC  & AUC   & Sim   & CC    & sAUC  & AUC            \\
\hline
{\tt ShallowNet} \cite{pan-cvpr-2016}       & 0.361 & 0.407 & 0.498 & 0.821 & 0.369 & 0.433 & 0.501 & 0.855          \\ %
{\tt ShallowNet+GRU}                        & 0.338 & 0.414 & 0.495 & 0.856 & 0.350 & 0.438 & 0.508 & 0.884          \\ %
\hline
{\tt C3D+Conv}                              & 0.347 & 0.399 & 0.643 & 0.860 & 0.445 & 0.561 & 0.663 & 0.907          \\
{\tt C3D+GRU}                               & 0.344 & 0.425 & 0.507 & 0.861 & 0.466 & 0.554 & 0.570 & 0.909          \\ %
\hline
{\tt RGP} (Ours)                            & \bf0.483 & \bf0.586 & \bf0.702 & \bf0.912 & \bf0.478 & \bf0.588 & \bf0.682 & \bf0.924 \\ %
\hline
\end{tabular}
\vspace{-6pt}
\caption{Evaluation of gaze prediction on the VAS and Hollywood 2 dataset.}
\label{tbl:results_gazepred}
\medskip
\small \centering
\begin{tabular}{ l|c|c|c|c|c}
\hline
Method & Random Uniform & Central Bias \cite{mathe-pami-2015} & SF+MF+CB \cite{mathe-pami-2015} & Human \cite{mathe-pami-2015} &  {\tt RGP} (Ours) \\ \hline
AUC & 0.500  & 0.840 & 0.871 & \textbf{0.936}  & \textbf{0.924}  \\
\hline
\end{tabular}
\vspace{-6pt}
\caption{Gaze prediction results in terms of AUC for Hollywood2 dataset.}
\label{tbl:results_gazepred_hollywood}
\vspace{-6pt}
\end{table*}

\begin{figure*}[t]
\centering
\includegraphics[width=0.99\textwidth]{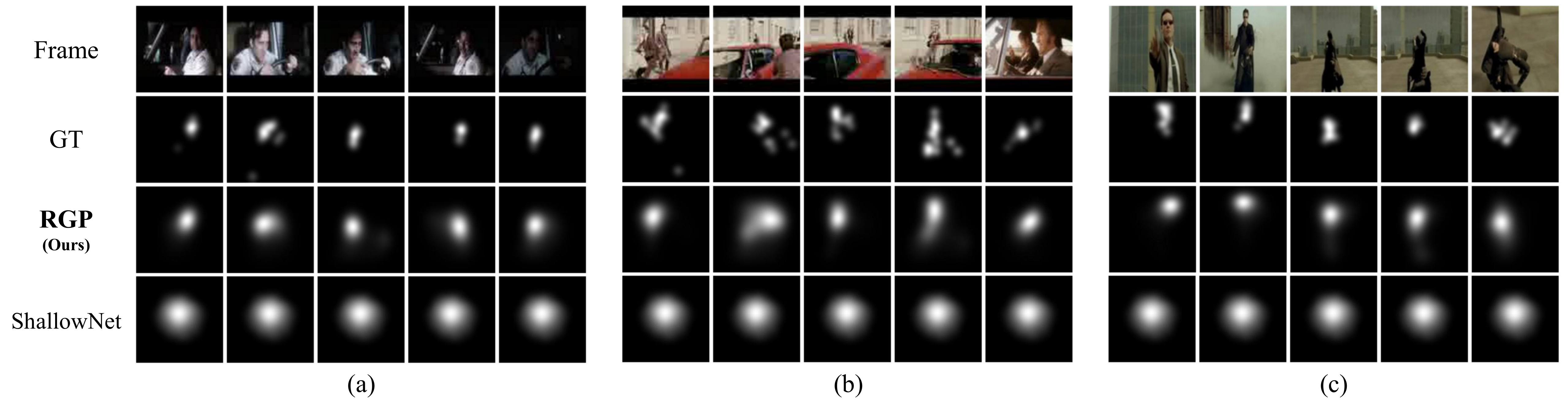}
\vspace{-3pt}
\caption{Examples of gaze prediction for video clips of Hollywood2 in (a--b) and VAS in (c). In each set,
we show five representative frames of the clip, along with GTs and predicted gaze maps predicted by different methods.
}
\vspace{-10pt}
\label{fig:examp_gazepred}
\end{figure*}

We evaluate gaze prediction performance by measuring similarities between the predicted and groundtruth (GT) gaze maps of test sets.
We follow the evaluation protocol of \cite{jiang-cvpr-2015,mathe-pami-2015,pan-cvpr-2016}.
Each algorithm %
predicts a ($49\times49$) gaze map for each frame, to which %
we apply Gaussian filtering with $\sigma=2.0$.
We then upsample it to the original frame size using bilinear interpolation.
The GT gaze map is obtained by averaging multiple subjects' fixation points, followed by a Gaussian filtering with $\sigma=19$.
After min-max normalization of predicted and GT gaze maps in a range of $[0,1]$,
we compute performance metrics averaged over all the frames of each test clip.
The performance measures include the similarity metric (Sim), linear correlation coefficient (CC),
shuffled AUC (sAUC) and Judd implementation of AUC (AUC),
whose details can be found in \cite{riche-iccv-2013}.
To compare with the results in \cite{mathe-pami-2015},
we follow the evaluation procedure of \cite{mathe-pami-2015};
we uniformly sample 10 sets of 3,000 frames from test video clips, and report averaged performance.

\textbf{Baselines}.
The ShallowNet~\cite{pan-cvpr-2016} is one of the state-of-the-art methods for saliency or fixation prediction.
Since it is designed for images not for videos, we test two different versions;
we separately apply it to individual frames, denoted by ({\tt ShallowNet}),
and integrate it with the GRU~\cite{cho-emnlp-2014} for sequence prediction, denoted by ({\tt ShallowNet+GRU}).
We also experiment two variants of our model to validate the effects of the recurrent component;
({\tt C3D+Conv}) is our ({\tt RGP}) excluding the GRU-RCN part,
and ({\tt C3D+GRU}) replaces the recurrent structure with vanilla GRU.

\textbf{Quantitative results}.
Table \ref{tbl:results_gazepred} reports gaze prediction results of multiple models on VAS and Hollywood2 EM datasets.
The variants of ShallowNets do not accurately capture human gaze sequences, and even with the recurrent model of ({\tt ShallowNet+GRU}).
Thanks to the representative power of the C3D motion feature and effectiveness of our recurrent model,
the proposed ({\tt RGP}) model significantly outperforms  all the baselines in all evaluation metrics with large margins.
Another advantage of the RGP model is that it needs relatively fewer parameters compared to other baselines,
being beneficial for integrating with video captioning models without a risk of overfitting.
Table \ref{tbl:results_gazepred_hollywood} compares our results with the best results of Hollywood2 reported in \cite{mathe-pami-2015} in terms of the AUC metric.
Our AUC of 0.924 is significantly higher than the best reported AUC of 0.871 in \cite{mathe-pami-2015}, only slightly worsen than the human level of 0.936.
For  VAS evaluation, we train models using the combined training set from VAS and Hollywood2, because the VAS dataset size is relatively small.
For Hollywood2 evaluation, we use Hollywood2 training data only to fairly compare with the results of \cite{mathe-pami-2015}.

\textbf{Qualitative results}.
Fig.\ref{fig:examp_gazepred} presents comparison of gaze prediction results between different methods and GTs on VAS and Hollywood2 datasets.
While the baselines, including ({\tt ShallowNet}) and ({\tt ShallowNet+GRU}),
do not correctly localize the gaze point with a bias toward the center.
On the other hand, our model can effectively localize gaze points over frame sequences.

\subsection{Evaluation of Video Captioning}
\label{subsec:results_vidcap}

\newcolumntype{P}[1]{>{\centering\arraybackslash}p{#1}}
\begin{table*}[t!]
\centering
\small
\begin{tabular}{l|P{.4cm}p{.4cm}p{.4cm}p{.4cm}p{.4cm}c|P{.4cm}p{.4cm}p{.4cm}p{.4cm}p{.4cm}c}
\hline

Dataset                                                                & \multicolumn{6}{c|}{VAS}                      & \multicolumn{6}{c}{LSMDC} \\ \hline
Language metrics                                                    & B1    & B2    & B3    & M     & R     & Cr    & B1    & B2    & B3    & M     & R     & Cr    \\ \hline
\multicolumn{13}{c}{\textbf{No spatial attention by gaze maps (\ie without RGP)}} \\ \hline
{\tt Temp-Attention} \cite{yao-iccv-2015}                         & 0.139 & 0.049 & 0.028 & 0.039 & 0.124 & 0.035 & 0.082 & 0.028 & 0.009 & 0.043 & 0.117 & 0.047 \\
{\tt S2VT+VGG16} \cite{venugopalan-iccv-2015}                     & 0.241 & 0.091 & \textbf{0.051} & 0.068 & 0.195 & 0.060 & 0.162 & 0.051 & 0.017 & \textbf{0.070} & 0.157 & \textbf{0.088} \\
{\tt S2VT+GNet} \cite{venugopalan-iccv-2015}                      & 0.233 & 0.088 & 0.043 & 0.069 & 0.189 & 0.058 & 0.142 & 0.041 & 0.015 & 0.065 & 0.153 & 0.083 \\
{\tt h-RNN+GNet+C3D} \cite{yu-cvpr-2016}                          & 0.255 & 0.099 & 0.038 & 0.067 & 0.181 & 0.055 & 0.128 & 0.038 & 0.011 & 0.066 & 0.156 & 0.070 \\ \hline
{\tt GEAN+GNet}                                                   & 0.259 & 0.102 & 0.041 & 0.068 & 0.196 & 0.057 & 0.154 & 0.050 & 0.016 & 0.067 & 0.153 & 0.091 \\
{\tt GEAN+GNet+C3D}                                               & 0.264 & 0.105 & 0.042 & 0.070 & 0.201 & 0.058 & 0.166 & 0.050 & 0.018 & 0.068 & 0.154 & 0.095 \\
{\tt GEAN+GNet+C3D+Scene}                                         & \textbf{0.274} & \textbf{0.118} & 0.046 & \textbf{0.075} & \textbf{0.211} & \textbf{0.080} & \textbf{0.166} & \textbf{0.050} & \textbf{0.018} & 0.069 & \textbf{0.157} & 0.084 \\ \hline

\multicolumn{13}{c}{\textbf{Spatial attention by RGP predicted gaze maps (\ie with RGP)}}                                                                                                                         \\ \hline
{\tt Temp-Attention} \cite{yao-iccv-2015}                         & 0.147 & 0.049 & 0.029 & 0.046 & 0.149 & 0.048 & 0.085 & 0.028 & 0.011 & 0.046 & 0.121 & 0.057 \\
{\tt S2VT+GNet} \cite{venugopalan-iccv-2015}                      & 0.268 & 0.101 & 0.044 & 0.073 & 0.199 & 0.069 & 0.131 & 0.038 & 0.013 & 0.066 & 0.153 & 0.080 \\
{\tt h-RNN+GNet+C3D} \cite{yu-cvpr-2016}                          & 0.273 & 0.101 & 0.045 & 0.073 & 0.196 & 0.073 & 0.146 & 0.046 & 0.017 & 0.067 & 0.151 & 0.074 \\ \hline
{\tt GEAN+GNet}                                                   & 0.282 & 0.119 & 0.049 & 0.077 & 0.209 & 0.075 & 0.152 & 0.051 & 0.016 & 0.068 & 0.152 & 0.081 \\
{\tt GEAN+GNet+C3D+Scene}                                         & \textbf{0.306} & \textbf{0.125} & \textbf{0.049} & \textbf{0.084} & \textbf{0.229} & \textbf{0.084} & \textbf{0.168} & \textbf{0.055} & \textbf{0.021} & \textbf{0.072} & \textbf{0.156} & \textbf{0.093} \\

\hline

\end{tabular}
\vspace{-3pt}
\caption{Evaluation of video captioning with or without the RGP model for VAS and LSMDC datasets.
For language metrics, we use BLEU (B), METEOR (M), ROUGE (R), and CIDEr (Cr), in all of which higher is better.}
\label{tbl:results_vidcap}
\bigskip

\small \centering
\begin{tabular}{ l|P{2.5cm}|P{2.2cm}|P{2.2cm}|P{2.2cm}|P{2.2cm}}
\hline
Dataset &  ({\tt GEAN}) w/ RGP  & Uniform & Random Gaze & Central Gaze & Peripheral Gaze  \\ \hline
LSMDC  & \textbf{0.072} 	& 0.069 & 0.056  	  & 0.061 		 & 0.057 			    \\
VAS 	& \textbf{0.084} & 0.075 & 0.062  	  & 0.073 		 & 0.068 			    \\
\hline
\end{tabular}
\vspace{-3pt}
\caption{METEOR score comparison between learned and various fixed gaze weights.
}
\vspace{-6pt}
\label{tbl:results_vidcap2}
\end{table*}

In previous section, we validate that the proposed gaze prediction achieves state-of-the-art performances.
Based on such dependably predicted gaze maps, we test how much they help improve attention-based captioning models.
For evaluation, each video captioning method predicts a sentence for a test video clip,
and we measure the performance by comparing between its prediction and the groundtruth sentence.
We use four different language similarity metrics,
BLEU \cite{Papineni-acl-2002}, METEOR \cite{Banerjee-acl-2005}, ROUGE \cite{Lin-was-2004} and CIDEr \cite{Vedantam-arxiv-2014}.

\textbf{Baselines}.
We compare with four state-of-the-art video captioning methods.
First, ({\tt Temp-Attention})~\cite{yao-iccv-2015} is one of the first soft temporal attention models for video captioning.
Second, the S2VT~\cite{venugopalan-iccv-2015} is a sequence-to-sequence model
that directly learns mappings between frame sequences to word sequences.
We test two variants denoted by ({\tt S2VT+VGG16}) and ({\tt S2VT+GNet}) according to frame representation
VGGNet-16 and GoogLeNet.
Finally, ({\tt h-RNN+GNet})~\cite{yu-cvpr-2016} is a hierarchical RNN model that also leverages a soft attention scheme to generate multiple sentences.
For ({\tt Temp-Attention}), we use the source code proposed by original authors. %
For ({\tt S2VT+*}), we transform the original Caffe code %
into TensorFlow, in order to integrate with the gaze prediction module.
We implement ({\tt h-RNN+*}) by ourselves because no code is available.

\newcolumntype{P}[1]{>{\centering\arraybackslash}p{#1}}
\begin{table*}[t]
\centering
\small

\begin{tabular}{P{2.5cm} |P{3cm}|P{3cm}|P{4cm}}
\hline

\small
({\tt GEAN}) w/ RGP vs  & ({\tt S2VT}) w/ RGP & ({\tt h-RNN}) w/ RGP & ({\tt Temp-Attention}) w/ RGP      \\
\hline
LSMDC 	           & \textbf{58.7} \% (176/300)  & \textbf{59.3} \% (178/300)  & \textbf{73.7} \% (221/300)    \\ %
\hline
VAS                & \textbf{61.0} \% (183/300)  & \textbf{69.7} \% (209/300)  & \textbf{76.7} \% (230/300)     \\ %
\hline

\end{tabular}

\vspace{-3pt}
\caption{The results of Amazon Mechanical Turk (AMT) pairwise preference tests on LSMDC and VAS datasets.
We present the percentages of responses that turkers vote for ({\tt GEAN}) w/ RGP against baselines with RGP.}
\label{tbl:results_AMT}
\end{table*}

\begin{table*}
\centering
\small
\begin{tabular}{P{1.25cm} |P{2.89cm}|P{2.89cm}|P{2.89cm}|P{2.89cm}}
\hline
\small
    & ({\tt GEAN})    & ({\tt S2VT}) & ({\tt h-RNN}) & ({\tt Temp-Attention})      \\
\hline
LSMDC        & \textbf{65.3} \% (196/300) & \textbf{58.0} \% (174/300)      &\textbf{59.7} \% (179/300)     & \textbf{60.7} \% (182/300) \\
\hline
VAS          & \textbf{67.0} \% (201/300) & \textbf{60.7} \% (182/300)      &\textbf{62.7} \% (188/300)     & \textbf{63.3} \% (190/300) \\
\hline

\end{tabular}
\vspace{-3pt}
\caption{The results of AMT pairwise preference tests between the models with or without RGP.
For example, the second column shows the percentages of Turkers' votes for ({\tt S2VT}) with RGP against ({\tt S2VT}) without RGP.}
\label{tbl:results_AMT_rgp}
\end{table*}

\begin{figure*}[t!]
\centering
\includegraphics[trim=0cm 4.8cm 0.1cm 0cm,clip,width=0.99\textwidth]{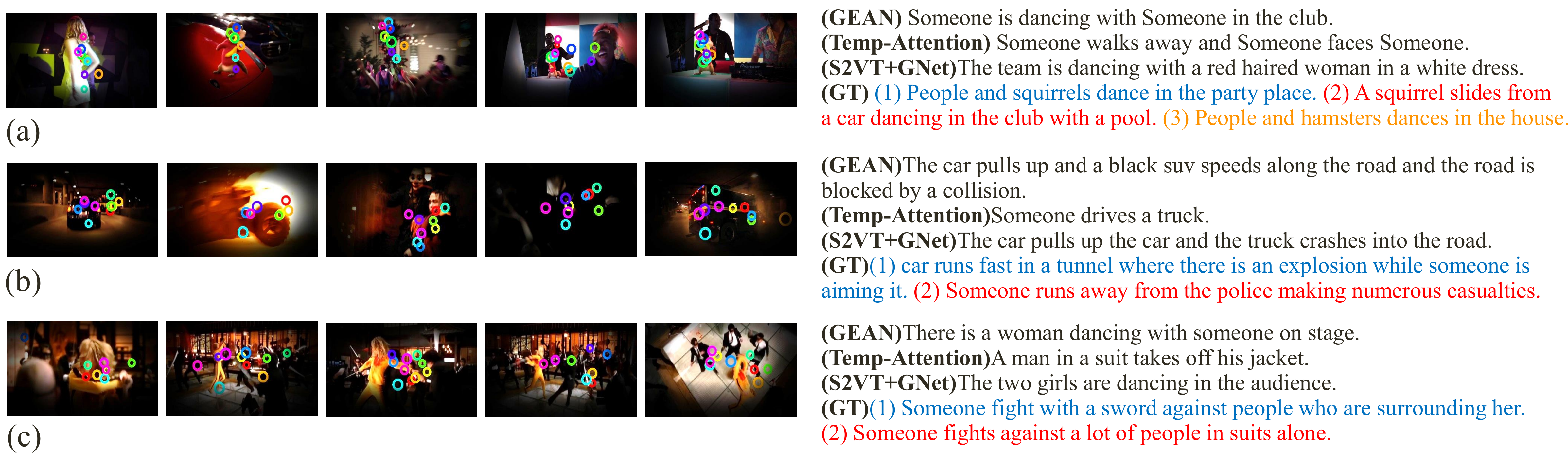}
\vspace{-3pt}
\caption{Video captioning examples of (a) correct, (b) relevant, and (c) incorrect descriptions.
In every frame, we present groundtruth (GT) human eye fixation with colored circles, and gaze prediction with white for attended regions.
We show captions predicted by different methods along with GTs. We present more, clearer, and larger examples in the supplementary.
}
\label{fig:examp_vidcap}
\end{figure*}

\textbf{Quantitative results}.
Table \ref{tbl:results_vidcap} shows quantitative results of different methods for video captioning.
We also run multiple variants of our GEAN model denoted by ({\tt GEAN+*}) according to different feature combinations.
We perform two sets of experiments with or without using the spatial attention by gaze maps that the RGP model predicts. %
The baselines without the RGP model means that they are executed as originally proposed.
For fair comparison, we use GoogLeNet inception5b layers as features for all baselines except ({\tt S2VT+VGG16}).
We obtain the results of ({\tt S2VT+VGG16}) for LSMDC dataset from the leaderboard of the LSMDC challenge.
Except this, we generate all the results by ourselves.

We summarize some experimental consequences as follows.
First, the proposed GEAN models achieve the best performance in each group of experiments for both datasets and with or without the RGP model.
Second, we observe that the performance of most methods increases with using spatial attention by gaze maps that the RGP predicts,
although the GEAN methods benefit the most from gaze prediction.
Such improvement is less significant in LSMDC than VAS dataset,
mainly because LSMDC has no gaze tracking data for training.
We remind that the RGP model is trained with VAS and Hollywood2 datasets.
Finally, experiments assure that it is the best for the GEAN model to use all the three visual feature pools,
as ({\tt GEAN+GNet+C3D+Scene}) attains the highest values in all the four groups of experiments.

\textbf{Effects of different gaze weights}.
Table \ref{tbl:results_vidcap2} compares captioning performance between different gaze weights within the RGP module.
For brief comparison, we report only METEOR scores. %
In the table, the performance with learned gazes by our model comes in the first column, and those of other baselines follow.
The uniform gaze assigns a uniform $1/49$ weight to $7 \times 7$ grid.
The random gaze selects a single bin randomly, while the central gaze picks the center $(4,4)$ bin in the grid.
Then, those one hot matrices of random and central gaze are smoothed by Gaussian filtering with $\sigma=1.0$.
Finally, the peripheral gaze is an $\ell_1$-normalized inverse of the central gaze.
As shown in Table \ref{tbl:results_vidcap2}, the learned gaze by our model leads the best captioning performance.
Among the fixed gaze weights, the uniform gaze is the best, which hints that it is better using the whole scene than attending on wrong parts of the scene.

\textbf{Qualitative results}.
Fig.\ref{fig:examp_vidcap} shows three examples of video captioning results for (a) correct description, (b) relevant description, and (c) incorrect description.
In frames, we present GT human eye fixation with colored circles, and gaze prediction with white for attended regions.
We also show the captions predicted by different methods along with GTs.
We observe that the spatial attention predicted by our method matches well with GT human eye fixation,
and description generated by our method are more accurate than the baselines.
We present more, clearer, and larger examples in the supplementary.

\subsection{Human Evaluation via AMT}
\label{subsec:results_human}

We perform user studies using Amazon Mechanical Turk (AMT) to observe general users' preferences on the generated descriptions.
We conduct pairwise comparison (A/B Test); in each AMT task, we show a clip and two captions generated by different methods in a random order, and ask turkers to pick a better one without knowing which comes from which methods.
For test cases, we randomly sample 100 examples each from LSMDC and VAS datasets.
We collect answers from three turkers for each test case.

Table \ref{tbl:results_AMT} shows the results of AMT tests on LSMDC and VAS datasets, in which we compare our ({\tt GEAN}) with the RGP model against the baselines with the RGP, including (\texttt{h-RNN}), ({\texttt{S2VT}), and (\texttt{Temp-Attention}).
We observe that general AMT turkers prefer output sentences of our approach to those of baselines.
Those response margins are more significant than language metric differences.

Table \ref{tbl:results_AMT_rgp} summarizes the results of AMT tests between the methods with or without RGP.
That is, for both our model and other baselines, we evaluate how much the gaze prediction by the RGP improves the caption qualities perceived by general users.
Consequently, even baselines with the RGP model obtains more votes than those without RGP.
It can be another evidence that gaze supervision helps even baselines to produce better descriptive sentences.

\section{Conclusion}
\label{sec:conclusion}

We proposed the Gaze Encoding Attention Network (GEAN) that leverage human gaze data to supervise attention-based video captioning.
With experiments and user studies on our newly collected VAS, caption-only LSMDC, and gaze-only Hollywood2 datasets,
we showed that multiple attention-based captioning methods benefit from  gaze information to attain better captioning quality.
We also demonstrated the GEAN model outperforms the state-of-the-art video captioning alternatives.

\medskip\noindent
\textbf{Acknowledgements}.
This research is partially supported by Convergence Research Center through National Research Foundation of Korea (2015R1A5A7037676).
Gunhee Kim is the corresponding author.

{\small
\bibliographystyle{ieee}
\bibliography{references}
}



\section*{Supplementary material}

# [Supplementary Material]
# Supervising Neural Attention Models for Video Captioning by Human Gaze Data

Youngjae Yu[†]   Jongwook Choi[†]   Yeonhwa Kim[‡]   Kyung Yoo[‡]   Sang-Hun Lee[‡]   Gunhee Kim[†]
[†]Department of Computer Science and Engineering   [‡]Department of Brain & Cognitive Sciences
Seoul National University, Seoul, Korea
yj.yu@vision.snu.ac.kr   {wookayin,billygma,yookyung20484,visionsl,gunhee}@snu.ac.kr

## Contents

## 1. More Experimental Results

### 1.1. Examples of Gaze Prediction

We present more examples of gaze prediction results in Figure 1 and 2. In each set, we present a sequence of frame images from video clips in VAS and Hollywood2 EM dataset. The corresponding groundtruth gaze map, and the gaze maps predicted by our RGP model as well as some baselines are presented. Additionally, we also show the result of ShallowNet [9], which is an image saliency network solely trained on the SALICON dataset [5] or on the Hollywood2 EM dataset.

### 1.2. Examples of Video Captioning

We present more examples of video captioning results in Figures 3 to 8. In Figure 3, gaze attention focused on facial expression and context information near main characters. In Figure 4, fovea region of RGP module shows that our model responds to an active object in motion. In contrast to (Temp-Attention) [6], GEAN generates a descriptive caption including major objects in motion and scene information. With RGP modules, attentive objects are preferentially selected. We show the original and masked frames on attended regions, and the generated captions by GEAN and baseline models.

### 1.3. Impact of different gaze prediction methods in video captioning

We below report the video captioning performance of our GEAN model, resulting from different gaze maps produced by other baselines (ShallowNet+GRU) and (C3D+GRU). The result is supplemental to main experiment. We observe that, the more accurate to human gaze, the better performance of video captioning. That is, our RGP model outperforms all gaze prediction baselines in terms of not only gaze prediction accuracies but also video captioning performance.

| Baseline Method | VAS | | LSMDC | |
|---|---|---|---|---|
| | METEOR | CIDEr | METEOR | CIDEr |
| Central Gaze | 0.073 | 0.064 | 0.061 | 0.063 |
| ShallowNet+GRU | 0.074 | 0.080 | 0.065 | 0.079 |
| C3D+GRU | 0.079 | 0.082 | 0.071 | 0.088 |
| RGP | **0.084** | **0.084** | **0.072** | **0.093** |

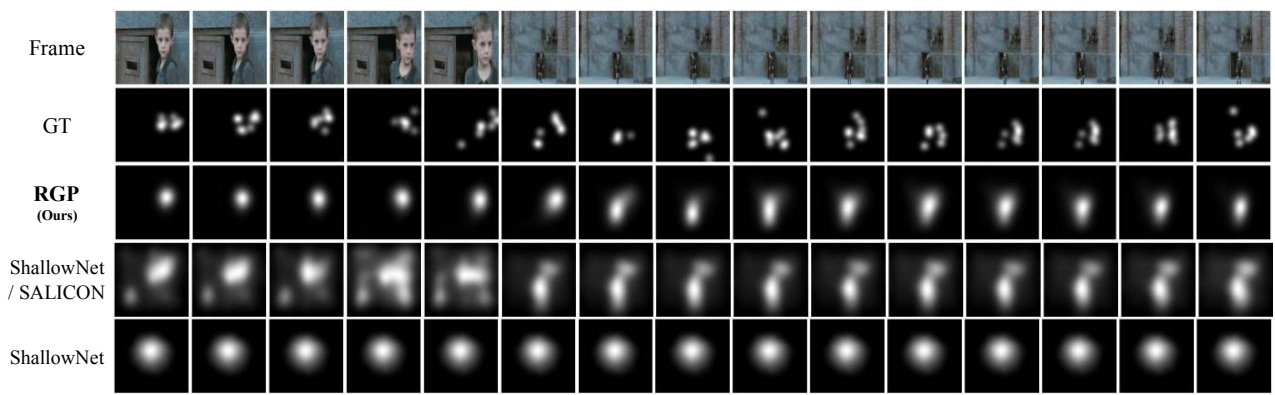

(a)

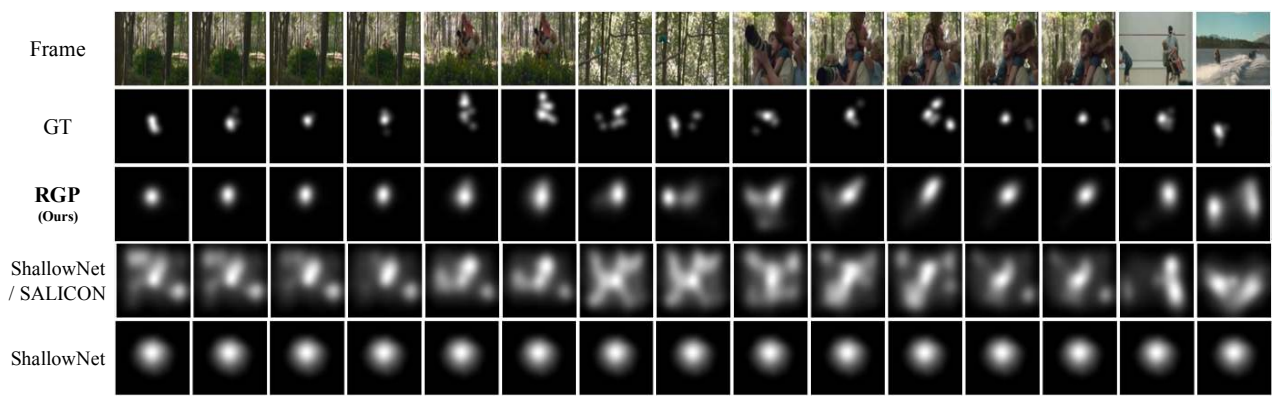

(b)

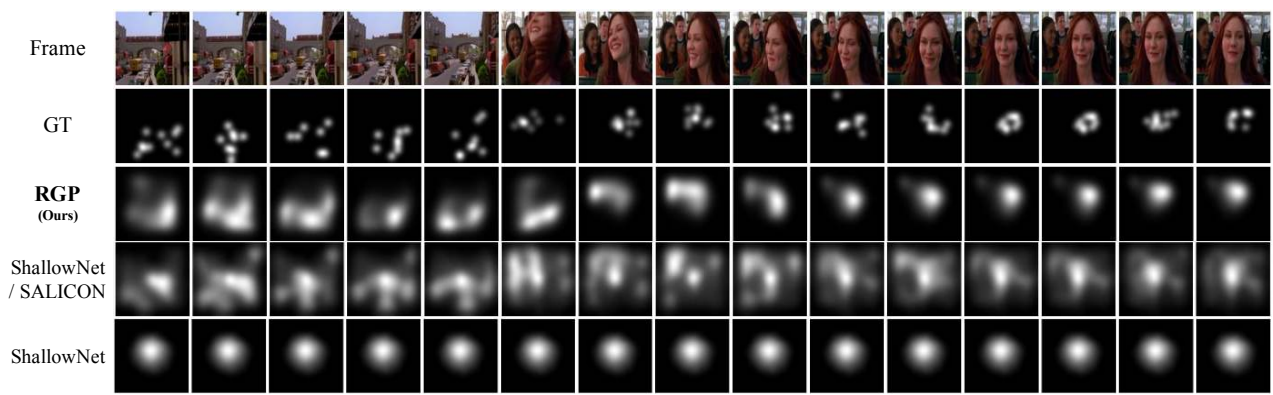

(c)

Figure 1: Examples of gaze prediction on the VAS dataset. In each set, we show equidistantly sampled (an interval of approximately 15 frames) frames from the video clip (the first row), the corresponding GT gaze map (the second row), and RGP's predicted gaze maps (the third row). We also present the results of ShallowNet trained on the SALICON dataset [5] (framewise, the fourth row) and on the training sets of Hollywood2 EM dataset (framewise, the fifth row).

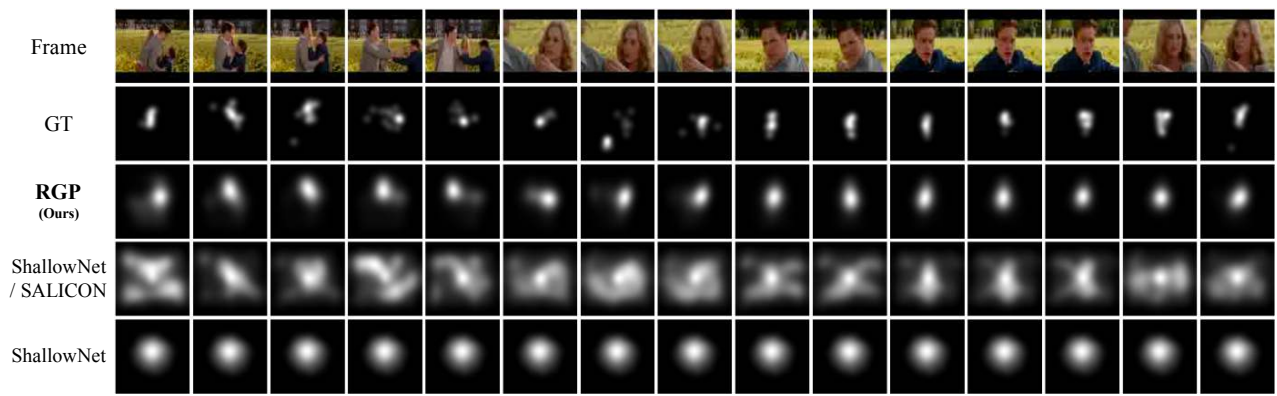

(a)

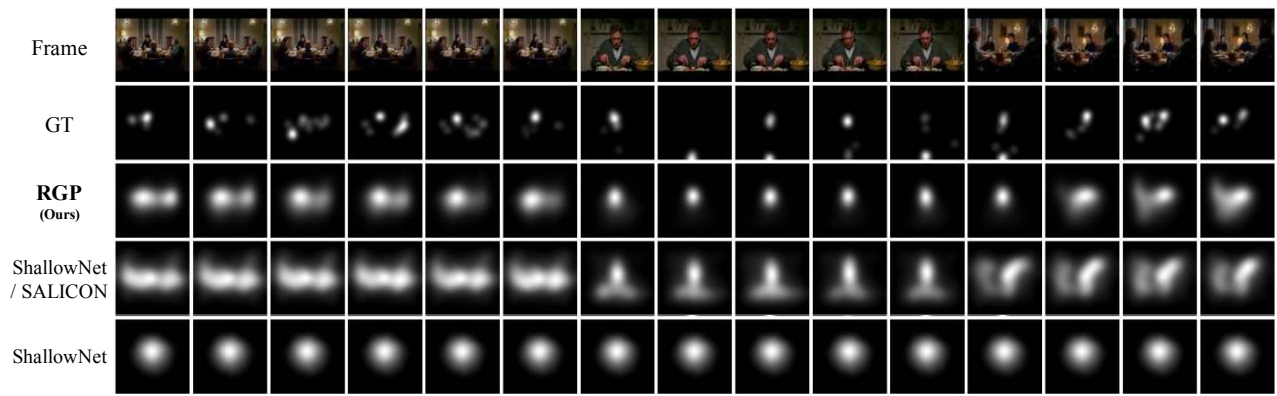

(b)

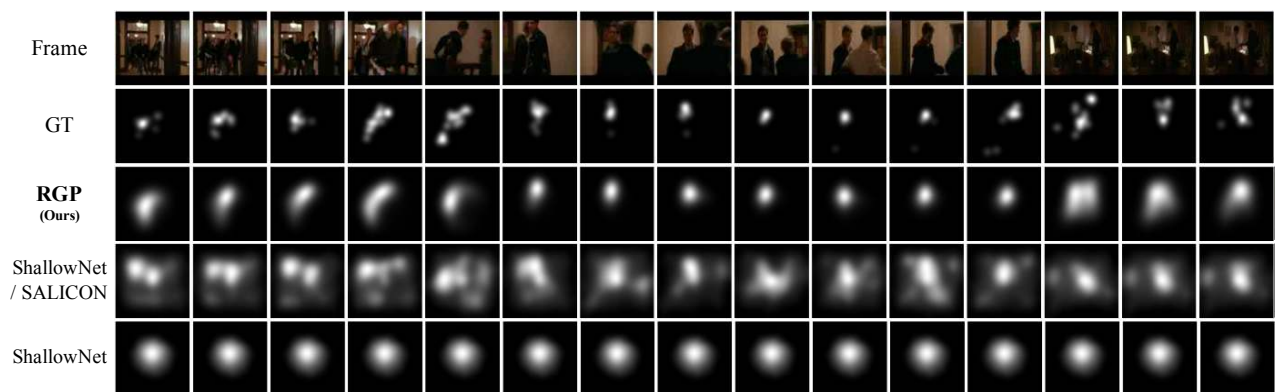

(c)

Figure 2: Examples of gaze prediction on the Hollywood2 EM dataset.

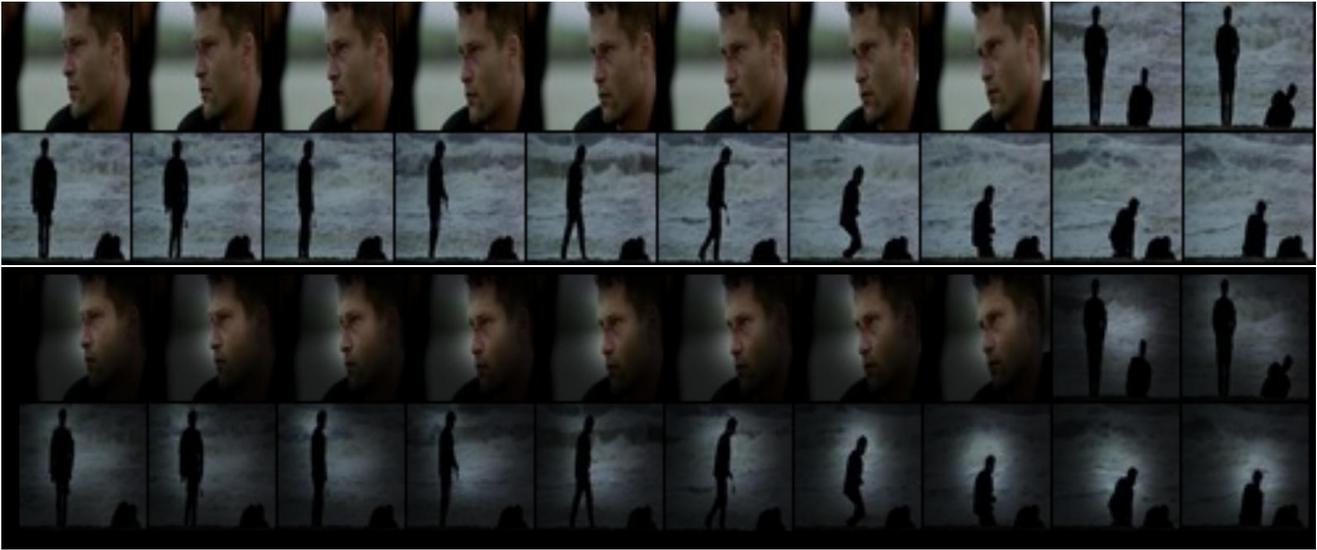

| | |
|---|---|
| (GroundTruth) | A man who are staring at somewhere suddenly falls down. |
| (GEAN) | SOMEONE looks up at the surf. |
| (S2VT) | SOMEONE looks up at the shore. |
| (Temp-Attention) | SOMEONE takes aim. |

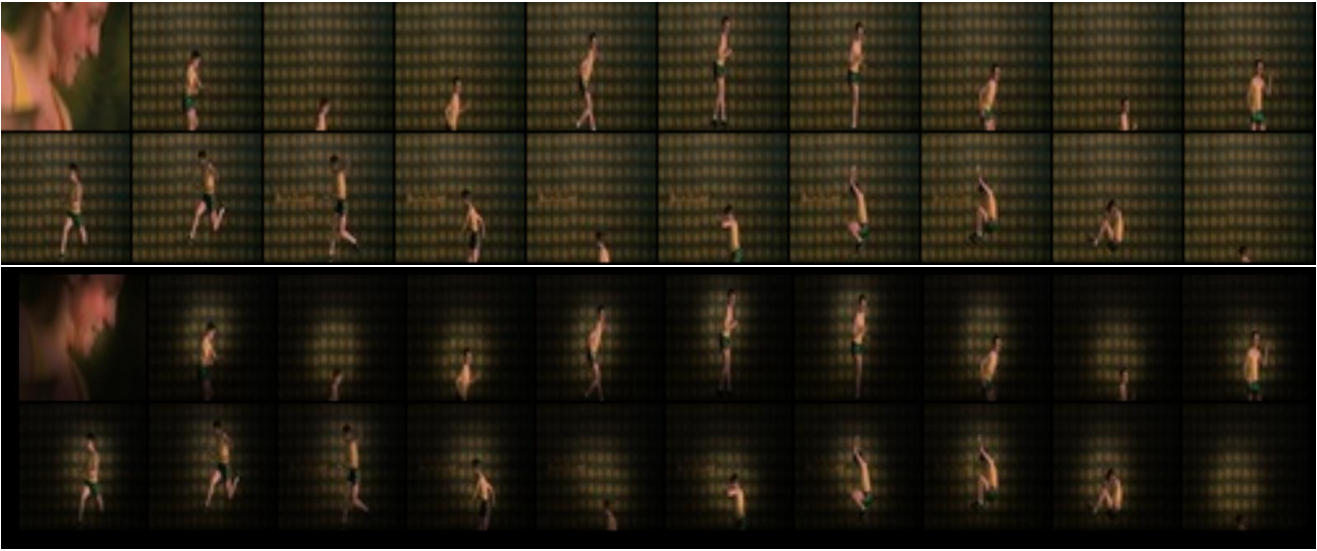

| | |
|---|---|
| (GroundTruth) | A boy jumps on the trampoline with changing his facial expression. |
| (GEAN) | SOMEONE looks at SOMEONE who is lying on the ground |
| (S2VT) | SOMEONE is wearing white dress. |
| (Temp-Attention) | SOMEONE looks down. |

Figure 3: Examples of video captioning results. A brighter (white) region in the middle row indicates the regions where the gaze of the model roughly attends to.

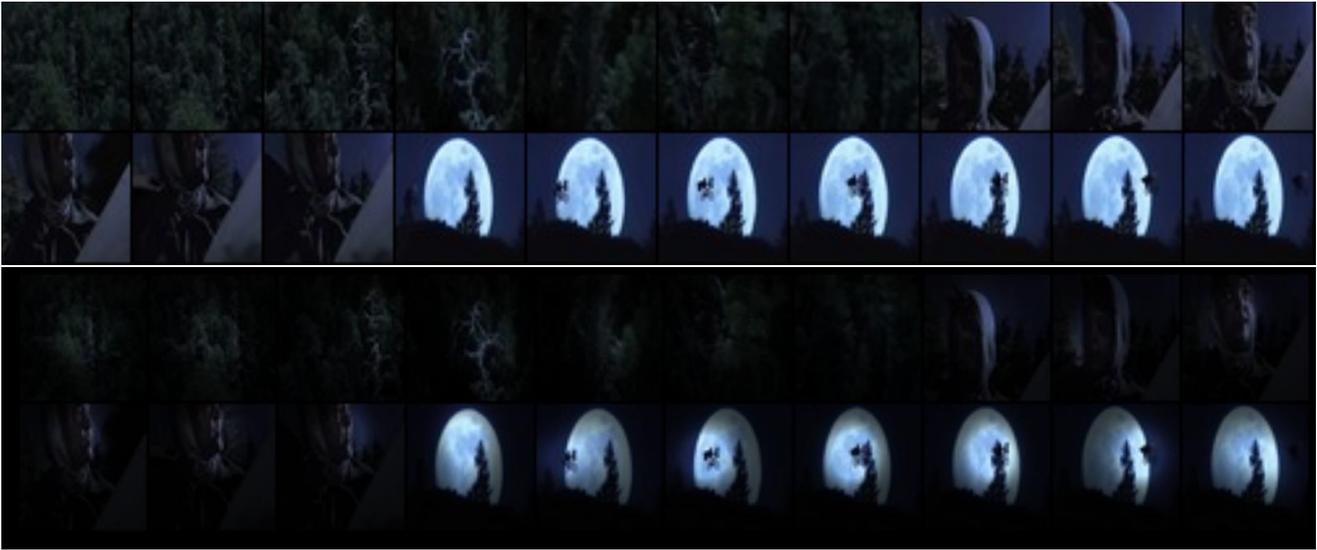

| (GroundTruth) | A little boy flies in the air by riding a bicycle. |
| (GEAN) | The bicycle flies out of the forest. |
| (S2VT) | There are many people are in the forest. |
| (Temp-Attention) | A man looks up. |

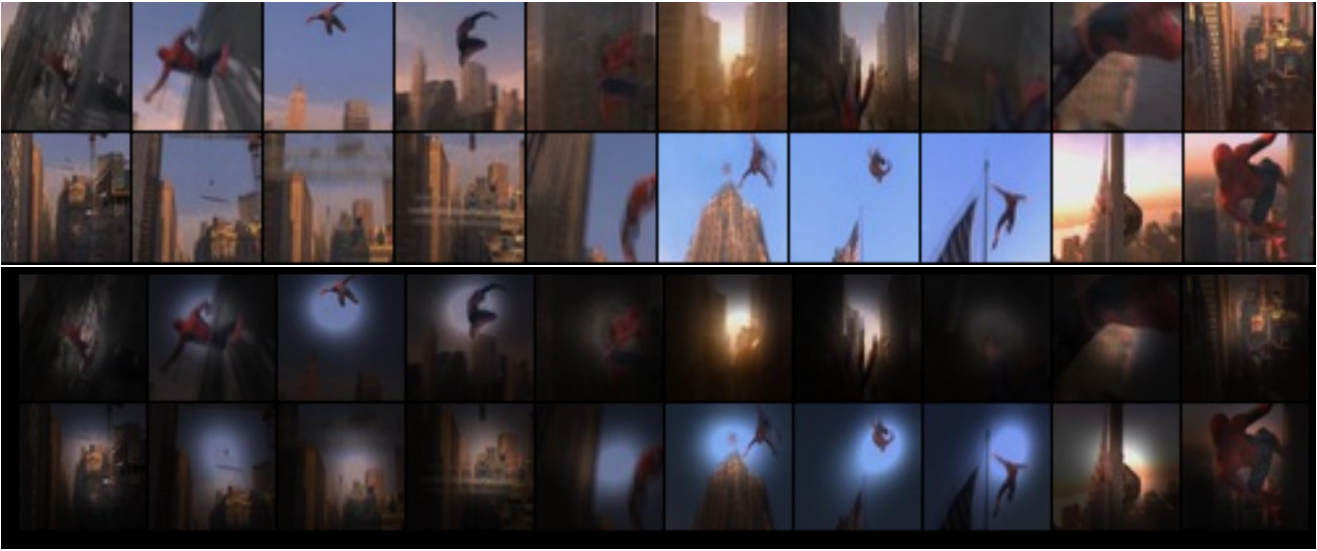

| (GroundTruth) | SOMEONE flies above the tall buildings, at last, climbs on the rooftop of the building. |
| (GEAN) | SOMEONE runs to the roof of the building and lands on the roof of the road. |
| (S2VT) | SOMEONE runs down the stair |
| (Temp-Attention) | SOMEONE takes a seat. |

Figure 4: Examples of video captioning results on VAS dataset.

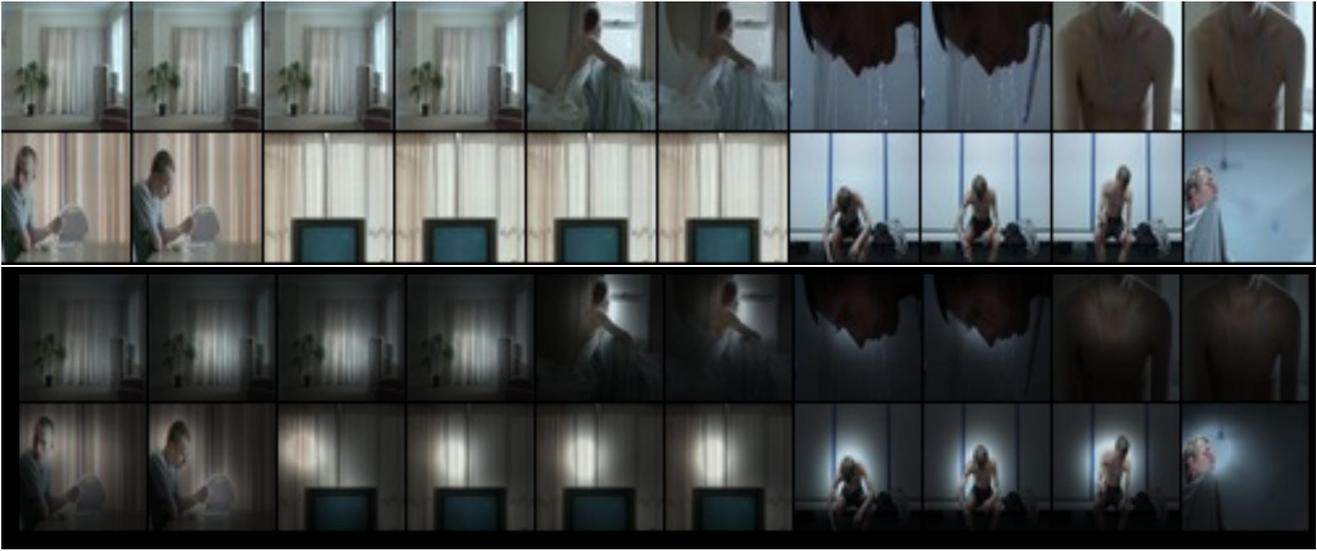

|                    |                                                                              |
|--------------------|------------------------------------------------------------------------------|
| (GroundTruth)      | SOMEONE's everyday lives like getting out of bed, going to swim and so on.   |
| (GEAN)             | SOMEONE is sitting on the bed in the hospital room.                          |
| (S2VT)             | SOMEONE is sitting on the bed in the kitchen                                 |
| (Temp-Attention)   | SOMEONE takes a shower.                                                      |

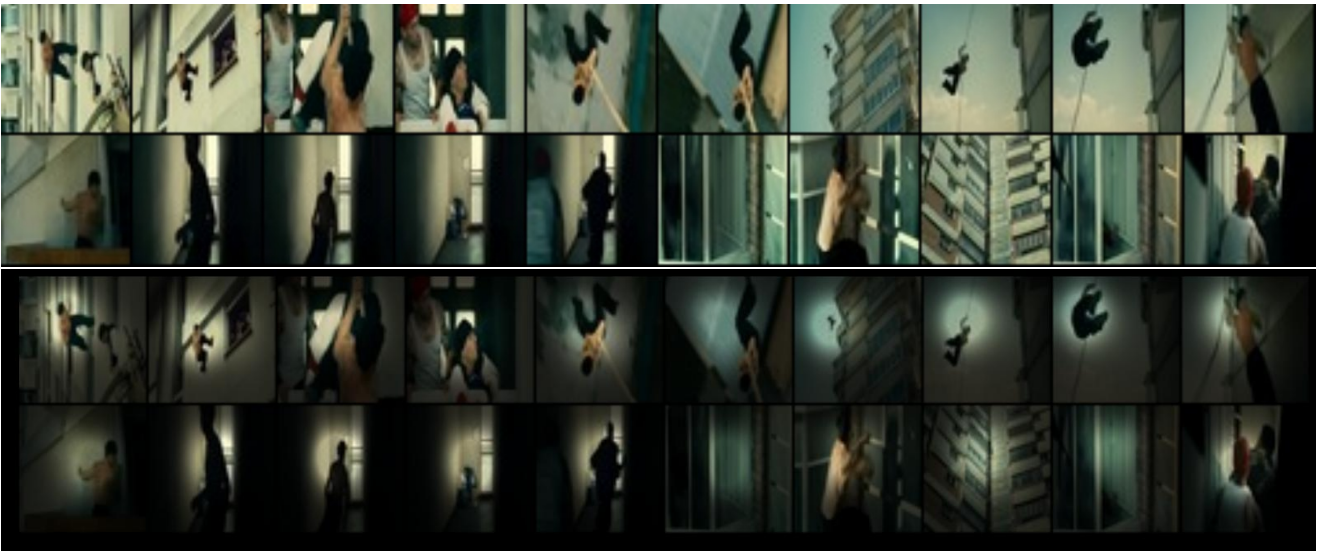

|                    |                                                                      |
|--------------------|----------------------------------------------------------------------|
| (GroundTruth)      | A man who is being chased by SOMEONE runs among the buildings.       |
| (GEAN)             | SOMEONE runs off and SOMEONE falls to the ground.                    |
| (S2VT)             | A man is runs to the stairs and jump up.                             |
| (Temp-Attention)   | SOMEONE is running.                                                  |

Figure 5: Examples of video captioning results on VAS dataset.

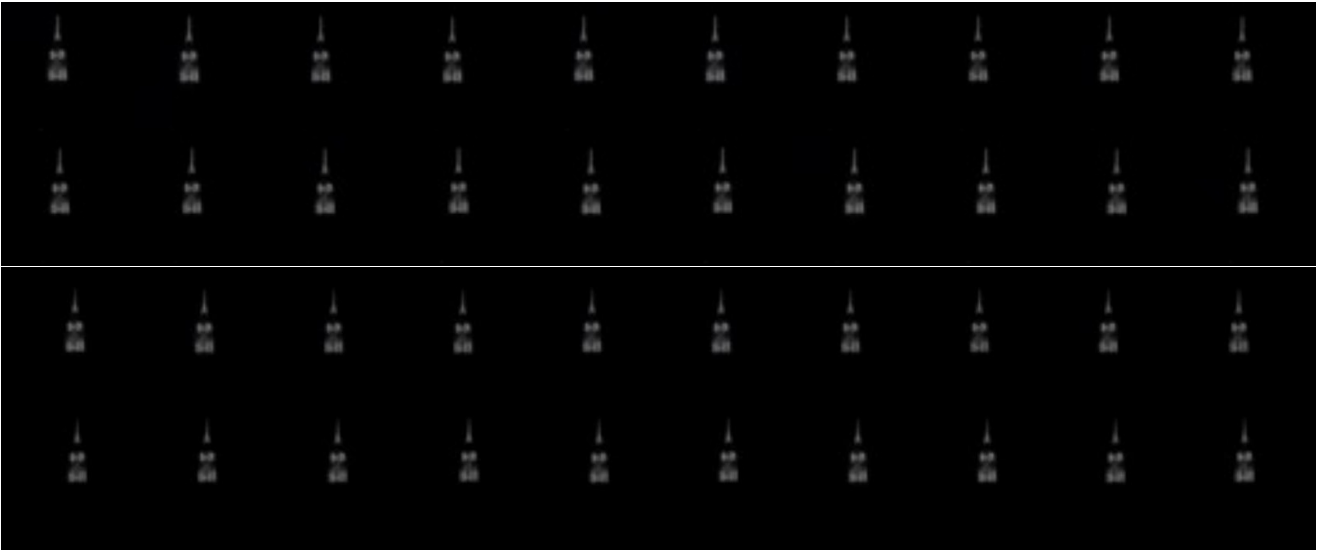

| | |
|---|---|
| (GroundTruth) | A kind of tower which is obscured with dim light in the night. |
| (GEAN) | The moon shines on the eiffel tower. |
| (S2VT) | The lights are glimming on the ground. |
| (Temp-Attention) | The lights dims. |

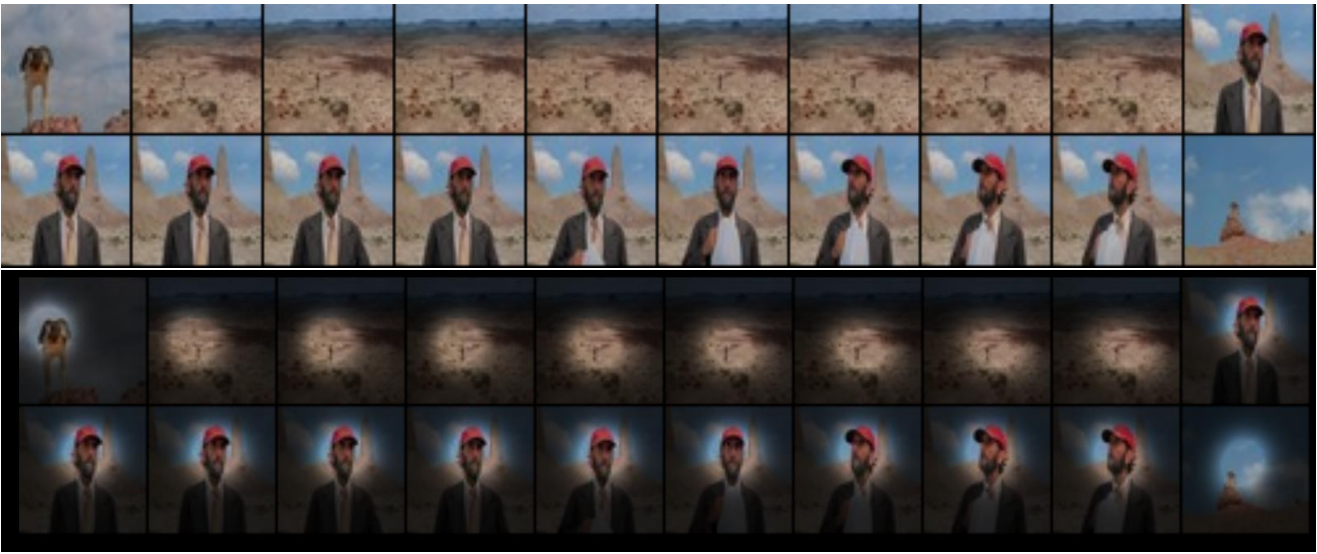

| | |
|---|---|
| (GroundTruth) | SOMEONE is wandering on the center of the desert. |
| (GEAN) | A man is walking along a field. |
| (S2VT) | A man is walking along a beach. |
| (Temp-Attention) | A man in suit standing. |

Figure 6: Examples of video captioning results on VAS dataset.

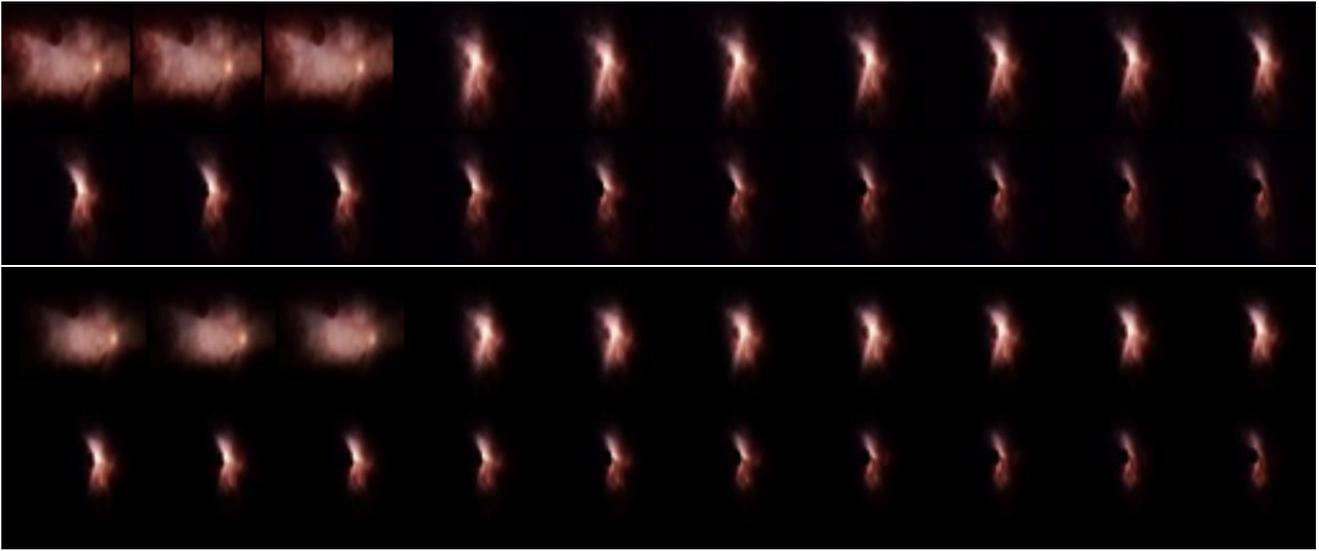

| | |
|---|---|
| (GroundTruth) | There is a total eclipse of the sun. |
| (GEAN) | The sun shines brightly in the night sky. |
| (S2VT) | There is a sphere in the dark light. |
| (Temp-Attention) | As the sun rises to the ground SOMEONE takes aim. |

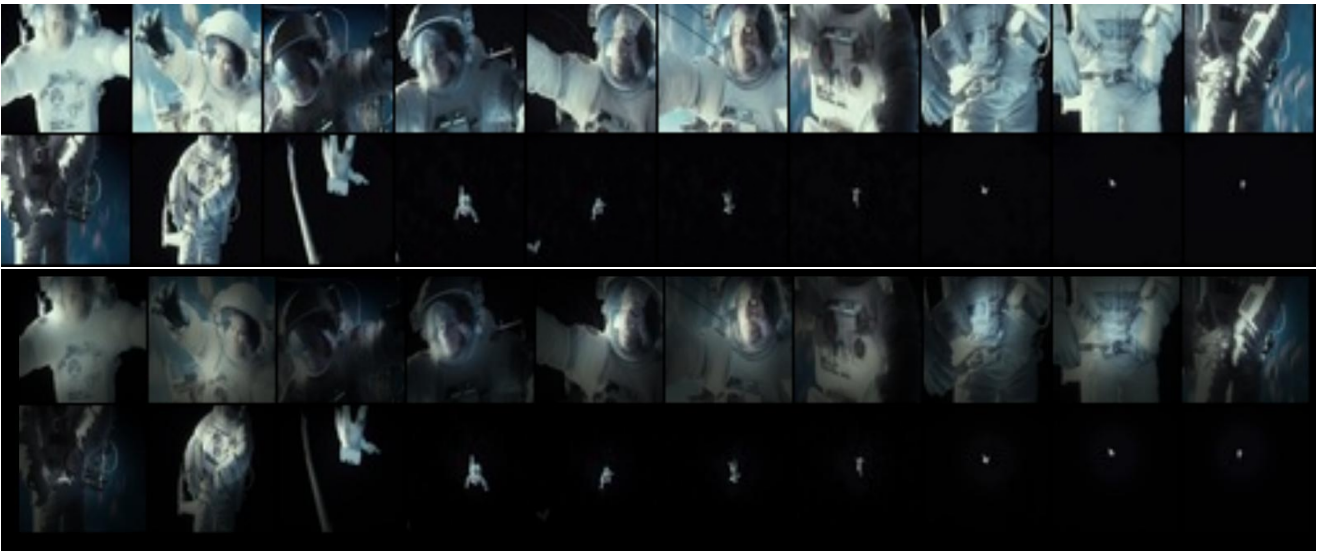

| | |
|---|---|
| (GroundTruth) | An astronaut flies away turning around with some debris from a meteors explosion. |
| (GEAN) | The man is thrown backwards in the air. |
| (S2VT) | A man is in a white suit. |
| (Temp-Attention) | A man is in the air. |

Figure 7: Examples of video captioning results on VAS dataset.

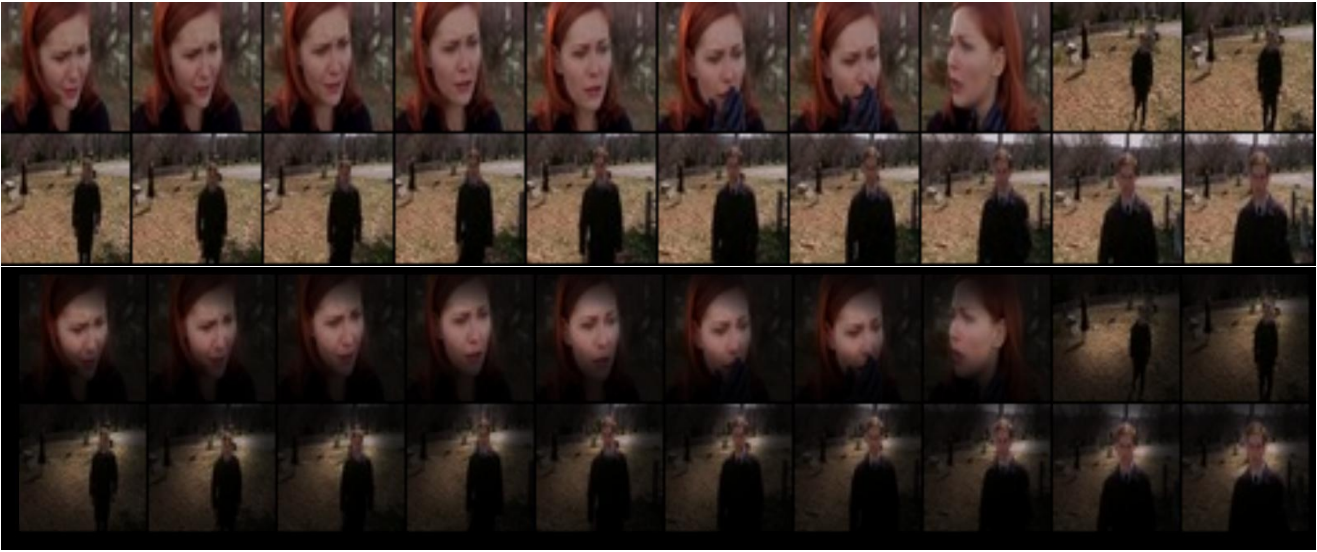

|                    |                                                                |
|-------------------:|----------------------------------------------------------------|
| (GroundTruth)      | At the cemetery, a woman cries and a man leaves from her.      |
| (GEAN)             | SOMEONE looks at SOMEONE.                                      |
| (S2VT)             | SOMEONE walks away.                                            |
| (Temp-Attention)   | She takes a deep breath.                                       |

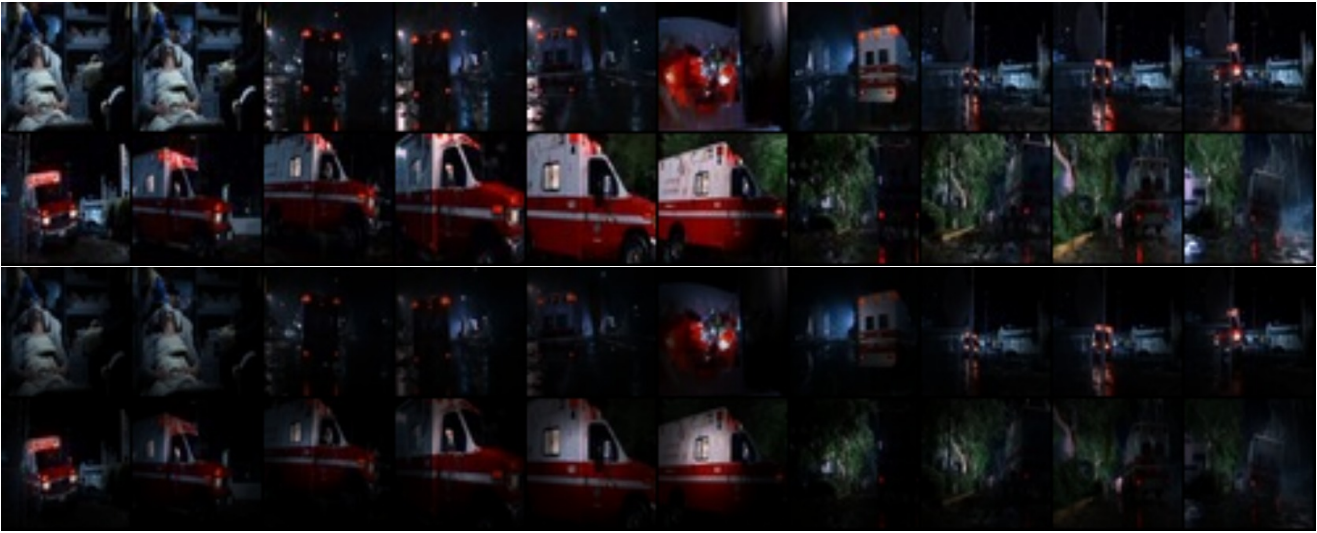

|                    |                                                                                |
|-------------------:|--------------------------------------------------------------------------------|
| (GroundTruth)      | Accidents happen in a row while an ambulance carrying an urgent patient runs.  |
| (GEAN)             | The car pulls up and a car swerves to avoid a collision.                       |
| (S2VT)             | the car pulls up the road and drives off.                                      |
| (Temp-Attention)   | Someone drives the truck.                                                      |

Figure 8: Examples of video captioning results on VAS dataset.

Table 1: Descriptive statistics of datasets

| Statistics / Datasets | M-VAD | MPII-MD | LSMDC | VAS | Hollywood2 |
|---|---|---|---|---|---|
| Number of sentences | 46,523 | 68,375 | 108,536 | 4,032 | – |
| Vocabulary size | 15,977 | 18,895 | 22,898 | 2,515 | – |
| Number of videos / images | 46,009 | 68,337 | 108,470 | 144 | 1,707 |
| Median length of sentence | 10 | 6 | 6 | 10 | – |
| Average term frequency | 31.0 | 34.7 | 46.8 | 16.2 | – |
| Fps (Frame per second) | – | – | – | 24 | 23-29 |
| Annotation | DVS | DVS | DVS | Caption, eye tracking | Action, eye tracking |
| Subjects (per video) | – | – | – | 31 (10–11) | 19 (19) |

## 2. Collection of VAS (Visual Attentive Script) Dataset

In this section, we describe the details of data collection. A descriptive statistics of the VAS dataset as well as other video datasets is given in Table 1.

**Criteria for video clip selection.** To build a library of video clips, we excerpted 144 visual scenes, each 15 seconds long, from various visual media sources including 116 films, 4 music videos and 4 television commercials. To ensure the video captioning algorithms get trained and validated by stimuli with sufficient diversity and external validity, we take into account the following factors when sampling the video clips. First, the referred media sources are diverse and balanced in terms of their genre, covering action, adventure, biography, comedy, documentary, drama, family, fantasy, mystery, horror, romance, sci-fi, sport, thriller, and western. Our film expert, manually selects those sources by referring to the Internet Movie Database (IMDb) and previous studies [8]. Second, the scenes depicted in the selected clips induce diverse visual stimuli. Specifically, one eighth of the entire clips (144/8 = 18 clips) belongs to one of the eight emotional categories defined by the circumplex model of human core affects. Third, we attempt to select the clip that carries a unique, compact 'event' or 'story', so that it can be readily summarized with a few sentences by human viewers. An example video clip is shown in Figure 12.

**Movie Familiarity.** We conducted a survey on whether participants had watched movie clip before. We suggest four options: *Saw and remember*, *Saw but forgot*, *Never seen* and *Do not know*. Excluding *Do not know*, we score each options as 3, 2, and 1, respectively. The mean average score 1.1987 implies that our selected movie clips are generally unfamiliar to the participants. This familiarity report therefore supports that the collected gaze information has few intervention of previous memory. Figure 9 briefly shows the user survey result. We plan to report the survey result for each participants along with the dataset.

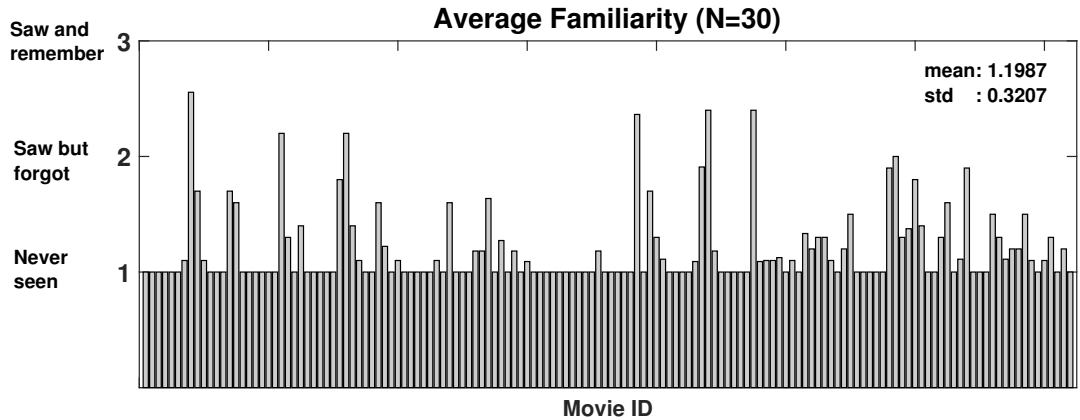

Figure 9: User survey of movie clip familiarity. Each bar represents the mean familiarity score of 30 participants, for each movie clip.

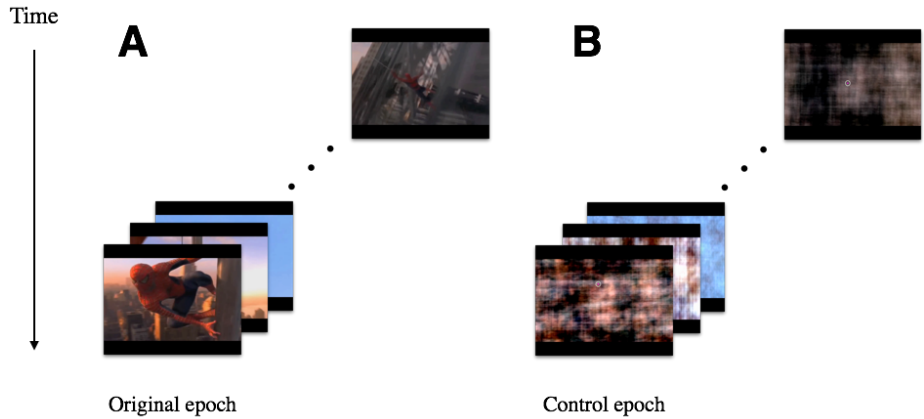

Figure 10: Example video clips used for eye-tracking. **A** is an example of the original clips. **B** is the control video clip corresponding the one shown in **A**. The bulls-eye marker was the target of fixation and moved around the display along the trajectory of the actual eye movements made by a participant when viewing the original clip shown in **A**.

**Viewing condition.** To avoid any unwanted differences in gaze activity due to different viewing conditions, all the excerpted clips are standardized to a fixed spatial dimension (1400-pixel width, 744-pixel height), a temporal frequency (24 frames per second), and a color mode (8-bit RGB). During gaze measurement, these standardized clips are resized to $1280 \times 680$ pixels and displayed on a LCD monitor (LG Flatron L1954TP with refresh rate of 60Hz and spatial resolution of $1280 \times 1024$ pixels) and viewed at a distance of 98 cm, subtending the visual angle of 21.83 °in width and 12.33 °in height. To ensure that gaze responses are determined solely by the visual information originated from the video clips, we removed the sounds completely from the original clips and then presented them in a quiet and light-controlled (reflected light or sound minimized) room built for audio-visual psychophysics experiments. Stimuli are displayed and controlled using MATLAB scripts in conjunction with Psychtoolbox-3.[1]

**Control video clips.** In addition to gaze, we simultaneously measure the size of the pupil to acquire the information about the arousal state of participants. However, since the dynamics of pupil size is majorly driven by the light and other low-level visual features [2], we regress out the variance in pupil size due to those nuisance factors. To this end, we create control clips that match the original clips in terms of spatiotemporal energy but are phase-scrambled such that any meaningful objects cannot be recognized (see Figure 10.**B** for an example). The specific steps for creating the control clips are as follows. First, an original clip is Fourier-transformed to give an amplitude spectrum and a phase spectrum. Second, random dot images of the same dimension as the original clip was also Fourier-transformed. Third, the amplitude spectrum of the original clip and the phase spectrum of the random dot clip are inverse-Fourier-transformed back into a hybrid clip in the image domain. This procedure is repeatedly applied to each and every original video clip, resulting in a total of 144 control video clips. By subtracting the pupil size of a control clip from that of its matching original, we can estimate the dynamics of pupil size driven by the endogenous factors including arousal levels [3, 4]. Thus our VAS dataset eliminates the effect of exogenous factors, such as luminance and saccade, and thereby measures the pupillary response to the movie context only. It is distinct from Hollywood2 EM gaze dataset [8] which provides raw pupil sizes. We encourage other researchers to use extra controled pupil size data for their research purposes.

## 2.1. Measuring the eye dynamics for VAS

**Participants.** Thirty participants (12 female and 19 male; average age is 22.08 ) with normal vision participated in the experiment and were all naive to the purpose of study. Each participant viewed 48 clips. As a result, each of the clips was viewed by a total of 10 different participants. Written informed consent was obtained from each participant, and the experiment were performed in compliance with the safety guidelines for human research, as approved by the Institutional Review Board of the authors' institution. All were naive to the purpose of study.

**Structure of data collection.** For each original clip, we collect a pair of eye dynamics datasets during two epochs of eye measurement, the first one acquired while participants viewing the original clip (original epoch) and then the second while

---
[1] http://psychtoolbox.org

viewing the matched control clip (control epoch). In the original epoch, participants watch the clip of 15-second length without performing any task while moving their eyes freely, with an instruction that they need to pay attention to the content (semantics in other words) of the clip. In the following control epoch, participants view the control clip with their eye dynamics being measured, but unlike in the original epoch, they perform a dynamic fixation task. To perform this task, participants had to move their gaze dynamically by fixating on a small bulls-eye (outer circle radius = 0.44 deg, inner dot radius = 0.10 deg) target that jumps and drifts along the spatio-temporal trajectory that mimicked the one actually made by themselves during the original epoch. In other words, the trajectory of the actual gaze during the original was measured and then 'replayed' in an online fashion during the control epoch. To ensure participants carry out the fixation task as instructed, we compute the correlation between the trajectories of the gaze made in the original and control epochs, and present the correlation value as a feedback at the end of each control epoch. Having measured the eye dynamics from both of the original and control epochs, we can further regress out any unwanted differences in pupil dynamics owing to gaze dynamics [7] between the original and control clips, helping us read out reliably endogenous cognitive factors (*e.g.* arousal levels) induced by the contents of the video clips. For each participant, we acquire a total of 48 pairs of eye data in a randomized order. Since ten participants viewed each of the 144 original clips and its paired control clip, we end up collecting 1,440 time-series, each 15-second long, epochs of eye dynamics (both in gaze and pupil size), respectively for the original and control clips.

**Apparatus and eye measurement setup.** We sample participants' binocular eye positions at 500 Hz by two infrared eye trackers: the EyeLink 1000 and the EyeLink 1000 plus (both Desktop Mount, SR Research; instrument noise, 0.01°RMS) are used in the main setup (see Figure 11) and in the auxiliary setup (not shown here), respectively. The two setups are identical to each other, and the only difference is that the chin rest is not needed in the auxiliary setup, which is capable of remote recording with a sticker put on the participant's forehead. The auxiliary setup is built to expedite the speed of data collection. The LED illuminator and camera are positioned side by side, at a distance of 74.5 cm from the observer, and angled toward the observer's face to insure that infrared light illuminates both eyes and is being reflected from both eyes and imaged on the camera sensor. An observer sits on a height-adjustable chair with his/her head supported by a forehead and chin rest (HeadSpot, UHCOTech), which are, together with the monitor, mounted on a height-adjustable table. For the second room (Eyelink 1000 plus), we do not use chin rest because it provides remote recording (sticker at forehead). To minimize body and head movements that compromise the quality of eye tracking measurements, we apply the following procedure: first, an observer is given enough time to find a comfortable arrangement of the chair, table, forehead, and chin rest by adjusting the heights of those devices. participants are then instructed to click a mouse button to indicate that they feel comfortable with a given setup and are ready to proceed to the next step.

**Calibration of the eye-tracker.** The eye tracker is calibrated using the built-in five point calibration routine (HV5), not only at the beginning of each daily session but also whenever the participant gets disengaged from a previously calibrated setup. During a session, the participant is allowed to take as many breaks as desired, disengaging from the eye tracking setup and moisturizing the eyes using disposable artificial tears as needed. We also let the participant take a mandatory 3-minute break each time after viewing 12 pairs of clips. Eye tracking signals are acquired in a 'pupil-corneal reflection (P-CR)' mode, and the pupil center is estimated using the ellipsoid fitting method, which is known to be robust to pupil occlusion by the eyelids. To detect any undesirable deviations from the current calibration, which would lead to inaccurate gaze measurements, we ask the participant to fixate on a centrally presented target for 4 seconds before each of the original measurement epoch. Once substantial deviations have occurred, we stop data collection and recalibrate the eye-tracker by adjusting the pupil or corneal reflection (CR) threshold.

**Preprocessing of eye dynamics data.** The EyeLink system estimates gaze position and pupil area using built-in proprietary software and provides those estimates to end users in a digitized format called 'EDF'. In this file format, gaze position estimates are in units of pixels of the stimulus monitor, and pupil area estimates are in arbitrary units. We analyze the data with the help of a publicly available script[2]. Because pupil information is unavailable during eye blinks, video-based gaze measurements can be contaminated by eye blinks. Thus, we identify eye blinks and associated time-series of gaze measurements and exclude them from subsequent analyses. Following Choe's work [1], eye blinks are judged to occur if any of the following two conditions are met: First, pupil data are missing for either eye; Second, pupil area measurements fluctuate abruptly with unrealistically large amplitudes ($> 50$ units per sample). As the data acquired immediately before and after ($\pm$ 200 ms) are likely to be contaminated by eye blinks, we remove and replace them with Not-a-Number (NaN) values. Pupil area values, which were originally provided in arbitrary units, are converted to units of 'percent change from the mean' to control the individual differences in absolute pupil size. To match the eye-tracking data to the video clips in terms of temporal resolution, we downsample the eye-tracking data from 500 Hz to 24 Hz.

---

[2]https://github.com/iandol/opticka/blob/master/communication/edfmex.m

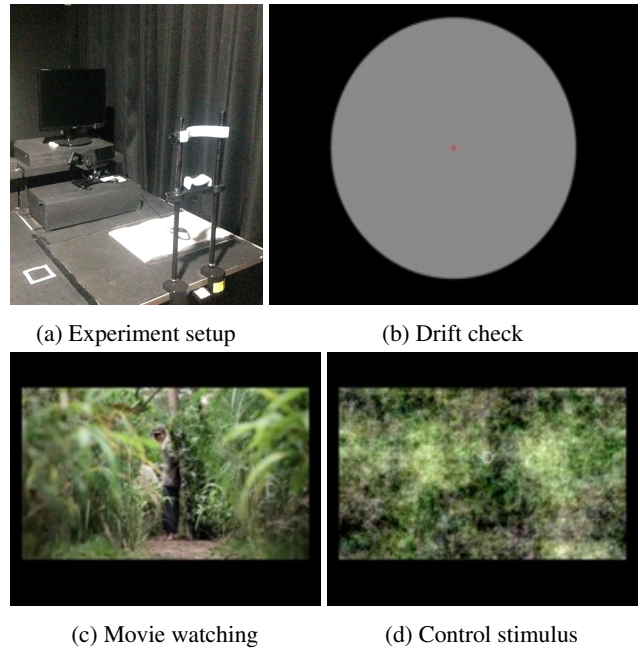

(a) Experiment setup      (b) Drift check

(c) Movie watching      (d) Control stimulus

Figure 11: (a) Experimental setup of eye measurement: the LCD monitor and the eye-tracker were placed on top of the height-adjustable table. Each trial of the collection of gaze tracking data consists of three steps, (b) drift check, (c) movie watching, and (d) control stimulus.

## 2.2. Acquiring the descriptive sentences for VAS

**Descriptive sentences acquired from participants.** We let each participant watch the 48 clips that they saw during the eye dynamics experiment, and ask them to provide three different types of descriptive summary, namely (i) overall summary, (ii) story summary, and (iii) characters and background summary. For overall summary, we gathers free understanding of summary for overall video clip. For the story summary, the participant generates a few sentences to reconstruct a story summarizing the core events occurred in the video clip (*e.g.* "A man puts headphones onto a woman. The woman turns around and hugs him"). For the characters and background summary, the participant indicates the main characters who lead that story (*e.g.* "A man and a woman") and the background where that story is played out (*e.g.* "In a club"). The participants provide those descriptive sentences through the Qualtrics online survey service (http://www.qualtrics.com) The participant may watch a given clip more than once if needed before completing the summaries, but are not allowed to edit their summaries or re-view the clip after completion.

**Descriptive sentences acquired from film experts.** To complement the summaries made by our participants, who were naive about movie films and not trained via official education for storytelling, we also acquire an additional set of descriptive sentences from two film experts who major in film directing and screenplay writing. They carry out the same tasks of generating the same three types of descriptive sentences for the entire set of VAS video clips. We assume that the experts with the eyes and brain trained for many years could offer the semantic descriptions of the video clips with the highest quality in a practical sense, which thus possibly can function as the goal that any video-captioning algorithms want to pursue.

Figure 12 is an snapshot visualization of gaze points in movie and corresponding descriptive sentences, where we show the descriptive sentences made by the participants who watched the video clip. The colors of the sentences match those of the circles in snapshots; The video clip shown here is excerpted from a motion picture, *Dark Knight*. The sentence made by our video caption algorithm (GEAN) is shown in parallel for comparison with the sentences made by human participants. Additionally, Figure 13 shows the frequency of words in the descriptive sentences. Descriptive sentences and gazed region has positive relationship. We also show more examples in Figure 14.

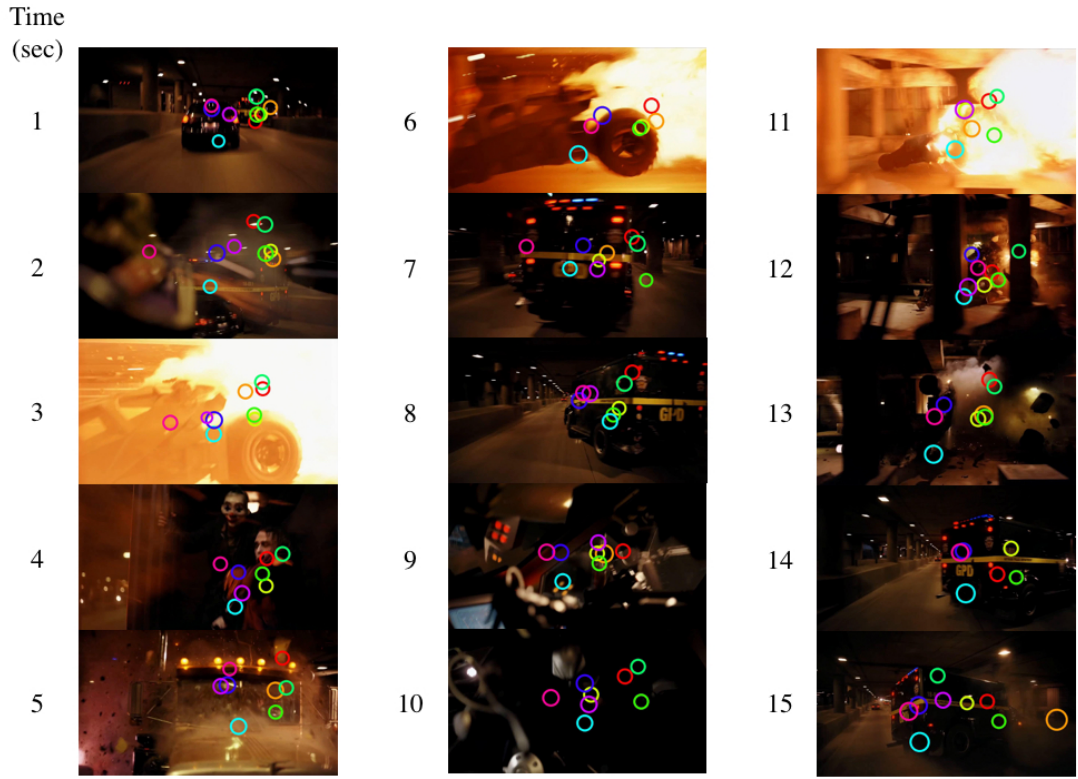

| Id | Story description |
|---|---|
| 1 | Batman drives his car but Joker hinders him to destroy. |
| 2 | Many cars engaged In a chase, one of them is shot then turned over. |
| 3 | A lot of cars are engaged in a chase then explode. |
| 4 | Batman drives his car and villains attack him by using arms like a mortar. |
| 5 | Batman on the Batcar chases some car while Jokers shoot something. |
| 6 | A criminal on the car fires missiles and Batcar is against that. |
| 7 | A car runs fast in a tunnel where there is an explosion while SOMEONE is aiming it. |
| 8 | Batman running through the town and Joker chasing him fight together. |
| 9 | A villain runs away from the police making numerous casualties. |
| GEAN | The car pulls up and a black suv speeds along the road and the road is blocked by a collision. |

Figure 12: The positions and sizes of the colored circles in each panel represent the gaze positions and pupil sizes, respectively, measured from the 9 individuals who watched this video clip and wrote description. Best viewed in color.

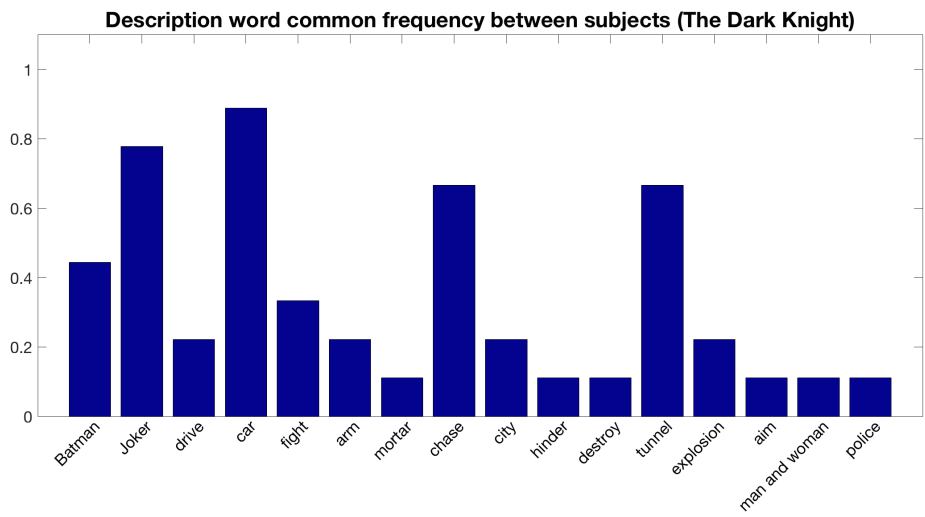

Figure 13: Frequency of words in the descriptive sentences. It was normalized to 0 to 1 scale. Frequently focused visual objects also frequently occur in description.

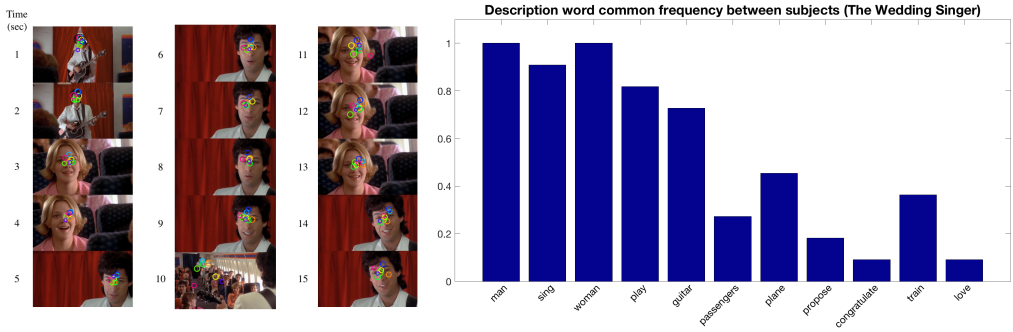

| Id | Story description |
|---|---|
| 1 | On the airplane, A man sings playing the guitar for some woman. |
| 2 | A man sings a song playing the guitar for a woman on the plane. |
| 3 | A man plays the guitar and sing a song for a woman. |
| 4 | A man performs for a woman. |
| GEAN | SOMEONE looks at SOMEONE who smiles in the plane. |

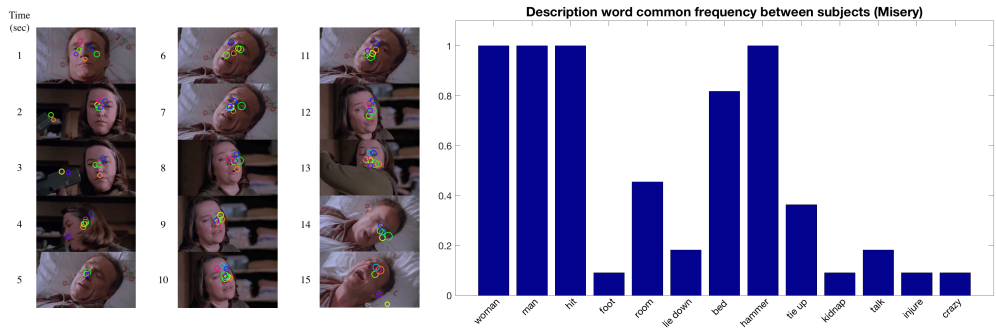

| Id | Story description |
|---|---|
| 1 | A man is bound on the bed, a woman breaks his leg by hitting with a hammer. |
| 2 | A man in middle age lying on the bed a woman breaks his leg by using a hammer. |
| 3 | A woman breaks a man's ankle by a hammer as the man lies on the bed. |
| 4 | A woman hits a man's leg by a hammer as the man is bound on the bed. |
| GEAN | SOMEONE looks at SOMEONE in a hospital. |

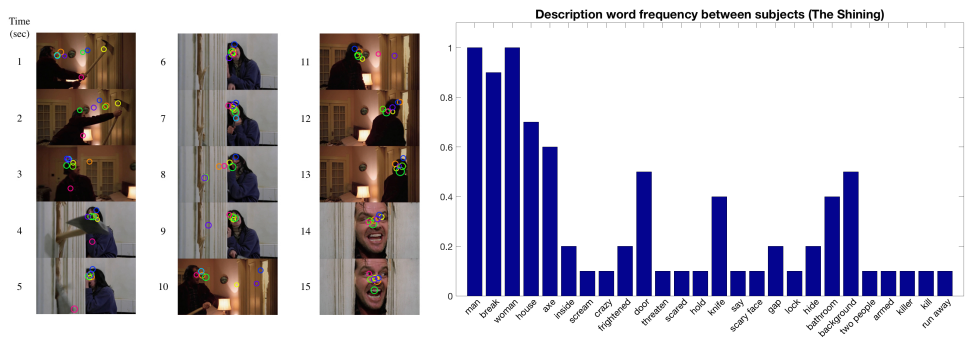

| Id | Story description |
|---|---|
| 1 | A man tries to enter the room by chopping the door with an ax. A frightened woman stands inside having a knife. |
| 2 | A man smashes the door with an ax and sticks his head out of the crack of the door. |
| 3 | A man tries to kill a woman with an ax. |
| 4 | A man beats SOMEONE badly for unknown reason. |
| GEAN | SOMEONE is in the bathroom and SOMEONE runs to the door. |

Figure 14: Examples for human gaze and description in the VAS dataset.


\end{document}

%% file: mymacros.tex
\usepackage{color}
\usepackage{amsmath}
\usepackage{amssymb}
\usepackage{multirow, varwidth}
\usepackage{verbatim}
\makeatletter
\newif\if@restonecol
\makeatother

\usepackage[ruled,vlined]{algorithm2e}
\usepackage{xr}
\usepackage[amsmath,thmmarks]{ntheorem}

\theorembodyfont{\normalfont}
\theoremseparator{.}

\theoremstyle{nonumberplain}

\newcommand{\bm}[1]{\boldsymbol{#1}}

\DeclareRobustCommand\onedot{\futurelet\@let@token\@onedot}
\def\onedot{. }
\def\eg{\emph{e.g}\onedot} 
\def\ie{\emph{i.e}\onedot}